\documentclass[12pt, draftclsnofoot, onecolumn]{IEEEtran}

\usepackage{bm}
\usepackage{amssymb}
\usepackage{nicefrac} 
\usepackage{graphicx}
\usepackage{booktabs}
\usepackage[top=0.75in, bottom=1 in, left=0.625in, right=0.625in]{geometry}

\usepackage{amsfonts}
\usepackage{amsmath}
\usepackage{amssymb}
\usepackage{pifont}
\usepackage{nicefrac} 
\usepackage{algorithm}
\usepackage{algorithmicx}
\usepackage{algpseudocode}
\usepackage{dsfont}
\usepackage{authblk}
\usepackage{bbm}
\usepackage{graphicx}
\usepackage{epstopdf}
\usepackage{subfigure}
\usepackage{stfloats}
\usepackage{eqnarray}
\usepackage{makecell}
\usepackage{multirow}
\usepackage{enumerate}
\usepackage{booktabs}
\usepackage{cite}
\usepackage{balance}
\usepackage{color}
\usepackage{bm}
\usepackage{caption}
\usepackage{mathtools}
\usepackage{url}

\newcommand{\sq}[1]{\Big[#1\Big]}

\newcommand{\one}{\mathbbm{1}}
\newcommand{\inp}[1]{\langle#1\rangle}

\newcommand{\E}[1]{\mathds{E}[#1]}
\newcommand{\EE}[1]{\mathds{E}\sq{#1}}

\newtheorem{theorem}{\textbf{Theorem}}

\newtheorem{lemma}{\textbf{Lemma}}
\newtheorem{condition}{\textbf{Condition}}

\newtheorem{remark}{\textbf{Remark}}

\begin{document}
	\title{Accelerating Wireless Federated Learning via Nesterov's Momentum and Distributed Principle Component Analysis}
	\author{\mbox{Yanjie Dong,~\IEEEmembership{Member, IEEE},}
		Luya Wang,
		Yuanfang Chi,~\IEEEmembership{Student Member, IEEE}, \mbox{Jia Wang,~\IEEEmembership{Member, IEEE}},
		\mbox{Haijun Zhang,~\IEEEmembership{Fellow, IEEE},}
		\mbox{Fei Richard Yu,~\IEEEmembership{Fellow, IEEE},}
		\mbox{Victor C. M. Leung,~\IEEEmembership{Life Fellow, IEEE}, Xiping Hu,~\IEEEmembership{Member, IEEE}}
%	\thanks{This work was supported by the National Nature Science Foundation (NSF) of China (62102266), 
%	the Pearl River Talent Recruitment Program of Guangdong Province (2019ZT08X603), 
%	Public Technology Platform of Shenzhen City (GGFW2018021118145859), 
%	Shenzhen Science and Technology Innovation Commission (R2020A045, KCXFZ20201221173411032), 
%	NSF of Beijing (L212004), 111 Project (B170003), 
%	the State Key Laboratory of Advanced Metallurgy (KF20-04), 
%	the Fundamental Research Funds for the Central Universities (FRF-TP-19-002C1, RC1631), 
%	and Beijing Top Discipline for Artificial Intelligent Science and Engineering, University of Science and Technology Beijing, 
%	and the Natural Science and Engineering Research Council of Canada 
%	(\emph{Corresponding author: Xiping Hu}).}
	\thanks{Y. Dong and X. Hu are with the Faculty of Engineering, Shenzhen MSU-BIT University, Shenzhen, China.}
	\thanks{L. Wang and J. Wang are with the College of Computer Science and Software Engineering, Shenzhen University, Shenzhen, China.}
	\thanks{H. Zhang is with Beijing Advanced Innovation Center for Materials Genome Engineering, 
		Beijing Engineering and Technology Research Center for Convergence Networks and Ubiquitous Services, 
		Institute of Artificial Intelligence, University of Science and Technology Beijing, Beijing, China.}
	\thanks{F. R. Yu is with the College of Computer Science and Software Engineering, Shenzhen University, Shenzhen, China and the Department of Systems and Computer Engineering, Carleton University, Ottawa, Canada.}
	\thanks{Y. Chi and V.~C.~M.~Leung is with the College of Computer Science and Software Engineering, Shenzhen University, Shenzhen, China, and the Department of Electrical and Computer Engineering, The University of British Columbia, Vancouver, Canada.}}
    \maketitle
    \begin{abstract}
    	%wireless federated learning, pca
  	A wireless federated learning system is investigated by allowing a server and workers to exchange uncoded information via orthogonal wireless channels.
  	Since the workers frequently upload local gradients to the server via bandwidth-limited channels, the uplink transmission from the workers to the server becomes a communication bottleneck.
  	Therefore, a one-shot distributed principle component analysis (PCA) is leveraged to reduce the dimension of uploaded gradients such that the communication bottleneck is relieved. 
  	A PCA-based wireless federated learning (PCA-WFL) algorithm and its accelerated version (i.e., PCA-AWFL) are proposed based on the low-dimensional gradients and the Nesterov's momentum. 
  	For the non-convex loss functions, a finite-time analysis is performed to quantify the impacts of system hyper-parameters on the convergence of the PCA-WFL and PCA-AWFL algorithms.
  	The PCA-AWFL algorithm is theoretically certified to converge faster than the PCA-WFL algorithm. 
  	Besides, the convergence rates of PCA-WFL and PCA-AWFL algorithms quantitatively reveal the linear speedup with respect to the number of workers over the vanilla gradient descent algorithm.
    Numerical results are used to demonstrate the improved convergence rates of the proposed PCA-WFL and PCA-AWFL algorithms over the benchmarks. 
    \end{abstract}

{\color{blue}
\section*{Nomenclature}
\begin{tabbing}
	\textbf{Scalars} \quad\:\:\= \textbf{Definitions} \\
	$h_{n,k}$ \> Channel coefficient vector of worker $n$ per frame $k$ \\
	$h_0$ \> Channel truncation threshold \\
	$d_0$ and $\hat d_0$ \> Dimension and reduced dimension of data sample \\
	$d_1$ \> Dimension of model parameter \\
	$c_{n,k}$ \> Inverse of channel scheduling probability of worker $n$ per frame $k$  \\
	$\eta$ \> Learning rate \\
	$L$ \> Lipschitz constant \\
	$\beta$ \> Momentum factor \\
	$K$ \> Number of frames \\
	$M$ \> Number of samples per worker \\
	$N$ \> Number of workers \\
	$\alpha$ \> Pathloss exponent \\
	$\rho_{n, k}$ \> Power control factor of worker $n$ per frame $k$ \\
	$\sigma^2$ \> Power of additive white Gaussian noise (AWGN)\\
	$c_1$ and $c_2$ \> Predetermined constants \\
	$\delta_{n,k}$ \> Propagation distance of worker $n$ per frame $k$ \\
	$p_0$ \> Transmit power budget of each worker\\
	$G$ \> Upper bound of gradient variance \\
	\textbf{Vectors} \\
	$z_k$ \> Accumulated noise per frame $k$ \\
	$u_k$ and $v_k$ \> Auxiliary variables per frame $k$\\
	$\one_{n,k}$ \> Scheduling policy of worker $n$ per frame $k$ \\
	$w_k$ \> Global model per frame $k$ \\
	$y_{n,k}$ \> Gradient of worker $n$ per frame $k$ \\
	$p_{n,k}$ \> Power control vector of worker $n$ per frame $k$\\
	$z_{n,k}$ \> Real part of AWGN  \\
	$\hat a_{m,n}$ \> The $m$th low-dimensional sample of worker $n$ \\
	$\nabla_{n,k}$ \> The $n$th intermittent gradient per frame $k$ \\
	$\hat\nabla_{n,k}$ \> The $n$th received gradient per frame $k$\\
	\textbf{Matrices}  \\
	$\hat U$ \> First $\hat d_0$ columns of global left-singular matrix \\
	$U$ and $V$ \> Global left- and right-singular matrices \\
	$\Lambda$ and $\Lambda_n$ \> Global and local principle component matrices \\
	$U_n$ and $V_n$ \> Local left- and right-singular matrices of worker  $n$ \\
	$A_n$ and $\hat A_n$ \> Local and low-dimensional datasets of worker $n$ \\
	\textbf{Functions} \\
	$f(\cdot)$ \> Empirical risk \\
	$f_n(\cdot)$ \> Local loss function of worker $n$
\end{tabbing}}

\section{Introduction}
%% wireless federated learning 
With the sheer volume of data at wireless edge terminals, the world starts to embrace machine intelligence applications at network edges (a.k.a., edge intelligence) \cite{9448012}.
As a popular framework of edge intelligence, a wireless federated learning (WFL) system can jointly infer knowledge from the proliferating data at edge terminals via the orchestration of cloud servers \cite{8976180}. 
By decoupling the learning process from the data collecting process, the WFL system can reduce the exposure of private information and provide immediate access to global models for the wireless terminals \cite{dongnetw}. 
Several emerging topics were investigated in the WFL system to improve the communication efficiency \cite{Chentobepublished, 9238427}, to enhance the robustness over passive and proactive adversaries \cite{yd_arxiv_byz, zhu2022bridging}, and to reduce the content access delay \cite{9252973, 9062302}.
For example, the lazy aggregation \cite{Chentobepublished} and quantized lazy aggregation \cite{9238427} are proposed to adaptively reduce the required communication rounds for training models in the WFL system. 
{\color{blue}
Since training deep neural networks becomes available for edge terminals, a federated learning-to-optimize framework was conceptualized for solving the NP-hard radio resource allocation problem by exploiting local data at all edge terminals \cite{Zappone2019}.}
	
While the aforementioned works are based on an \mbox{error-free} WFL system, the impacts of noise and fadings of wireless channels remain under-exploited on the energy consumption, communication efficiency, and convergence rate of the WFL system \cite{8970161}.
In this work, we aim at: 1) designing supervised learning algorithms for the WFL system (or WFL algorithms for brevity) to adapt to the wireless channels; and, 2) quantifying the impacts of the noise and fadings on the convergence rates of our proposed WFL algorithms. 

\textbf{Related works and motivations.}~When the federated learning system is deeply coupled with wireless communications, the success of WFL algorithms requires to adapting to the wireless channels and to suitably using the scarce radio resources \cite{8970161}. 
There are two concurrent research directions for the WFL system, i.e., the digital WFL system \cite{9435350, 9237168} and the analog WFL system \cite{9042352, 9749966}.
Since the digital WFL system uses the digital modulations to provide reliable communication tunnels, the joint design of WFL algorithms and radio resource management algorithms is required to efficiently utilize the radio resources for training.
More specifically, in the digital WFL system, radio resource allocation algorithms were designed to optimize the training loss \cite{9210812, 9562748}, training efficiency \cite{9827589, 9252924}, training duration \cite{9390199, 9760232, 9484767}, consumed energy \cite{9264742}, the number of correctly received model parameters per unit duration \cite{Yang2022} as well as the tradeoff between training duration and consumed energy \cite{Hou2023}. 
For example, the worker selection and bandwidth were jointly allocated to minimize the training loss in a hierarchical digital WFL system in \cite{9562748}.
A joint resource allocation of the minibatch size, uplink and downlink duration of workers was proposed to maximize the learning efficiency of the digital WFL system \cite{9252924}. 
By opportunistically scheduling workers, the number of selected workers and the training duration could be respectively optimized in  \cite{9237168} and \cite{9760232}  for the digital WFL system. 
{\color{blue}
Besides, the interplay between the reconfigurable intelligent surfaces (RISs) and the federated learning  was investigated to enhance the  spectrum-sensing accuracy and reliability of the training process by jointly allocating RIS association, wireless bandwidth, and phase-shifts of RISs \cite{Yang2022}.
The covert digital WFL system was investigated by jamming the potential eavesdroppers while striking a balance between training accuracy and energy consumption \cite{Hou2023}.}
The gradient quantization codebook was designed in \cite{9050465}  for the digital WFL system. 
Note that the design of radio resource management algorithms for computation-limited workers (e.g.,  tiny machine learning workers and sensors) could be a challenging task. 

Without requiring complex radio resource management, the over-the-air computing (OAC) technology allows the analog WFL system to aggregate the analog modulated local gradients by exploiting the waveform superposition property. 
In this vein, the analog WFL algorithms were proposed to adapt the information exchange to the AWGN channels \cite{9042352, 9459539} and fading channels \cite{9252950, 8870236, 9272666, 9014530, 9562537, 9076343, shao2021federated}.
The signal encoding algorithms were designed for the analog WFL system \cite{8807380, 8952884, 9272666}.  
For example, the channel noise is exploited to preserve the privacy of workers  in the analog WFL system \cite{9252950}.
The uncoded and coded transmission were respectively exploited in \cite{8870236} and \cite{9272666} to develop analog WFL algorithms, where the received strength of local gradients is limited by the farthest worker. 
The proposed analog WFL algorithm in \cite{9042352} averages the local gradients by assuming the same variance of local gradients, and recovers the average gradient via an approximate message passing mechanism. 
Besides, the momentum term is used to accelerate the analog WFL algorithm for (strongly) convex loss functions \cite{9562537}.
{\color{blue}
In order to achieve the drastic reduction of required bandwidth via the OAC technology, the aforementioned analog WFL algorithms need to obtain the unbiased estimators of the global model and global gradient by strong assumptions: 1) the independent and identically-distributed (i.i.d.) local gradients \cite{9042352}; or 2) i.i.d. locations of workers \cite{9459539, 9272666, 8870236, 9014530, 9252950, 9562537, 9076343}. 
Besides, the \mbox{finite-time} analysis was only performed for (strongly) convex loss functions \cite{9042352, 9459539, 9252950, 9562537, 9076343}.
Motivated by these facts, we envision that the potential application scenarios of the analog WFL system can be increased by removing the i.i.d. assumptions on local gradients and workers while calculating the global model and the global gradient. }

%% contributions
\textbf{Contributions.} Different from \cite{9042352, 9459539, 9272666, 8870236, 9014530, 9252950, 9562537, 9076343}, we consider the non-i.i.d. local gradients and the non-i.i.d. locations of workers. 
We aim at developing and analyzing the analog WFL algorithms for the non-convex loss functions. 
For the ease of implementation, we allow the workers to upload local gradients to the server via uncoded pulse amplitude modulation. 
When the workers have non-i.i.d. locations, the propagation distances between the workers and server are in high variance. 
Since the OAC technology has a limited degree of freedom for power control, the received strength of local gradients is confined by the farthest worker. 
Otherwise, the workers that are close to the server can scale their local gradients with larger factors  than workers farther away (i.e., the near-far effect of co-channel transmission) \cite{Goldsmith2005}. 
To remove the limitations of non-i.i.d. location of workers, the orthogonal channel usages\footnote{Since each frame consists of multiple resource elements that are temporally and spectrally orthogonal, a \emph{channel usage} indicates a utilization of each temporally-and-spectrally orthogonal resource element in the frame.} are allowed at workers. 
Due to the scarcity of radio resources, we leverage a one-shot distributed principle component analysis (PCA) algorithm \cite{liang2014improved} to reduce the dimension of samples and thereby the dimension of local gradients. 
The detailed contributions of our work are summarized as follows.
\begin{itemize}
\item We aim at developing analog WFL algorithms that can handle the near-far effect of workers and inherit the simplicity of uncoded pulse amplitude modulation. 
We leverage a one-shot distributed PCA algorithm \cite{liang2014improved} to reduce the dimension of samples and thereby the required radio resource. 
\item We introduce a truncated channel inversion to construct an unbiased estimator of the global gradient when the workers have non-i.i.d. locations.   
Based on the introduced truncated channel inversion and the one-shot distributed PCA, we start by proposing PCA-WFL algorithm for the non-convex loss functions. 
Motivated by the potential speedup capability of Nesterov's momentum, we propose a PCA-based accelerated WFL (PCA-AWFL) algorithm. 
Leveraging the intelligence networking \cite{9448012}, we also give an intuition on the potential speedup capability of Nesterov's momentum. 
\item Our theoretical results reveal that the \mbox{PCA-AWFL} algorithm converges with a faster rate than the PCA-WFL algorithm at an expense of a larger error-floor term. 
Besides, our theoretical results also confirm that the PCA-WFL and PCA-AWFL algorithms can benefit from the computational parallelism of numerous workers.
\end{itemize}

Several numerical tests are performed to verify the performance of our proposed \mbox{PCA-WFL} and \mbox{PCA-AWFL} algorithms on the \mbox{FMNIST} dataset \cite{xiao2017fashion}, CIFAR-10 dataset \cite{Krizhevsky2009}, and AWE dataset \cite{Zhang2021}.

\textbf{Organization and notations.} 
The remaining work is organized as follows. 
Several preliminaries are introduced in Section II.
The system model and problem description are provided in Section III. 
The algorithm design and corresponding finite-time analysis are established in Section IV.
Numerical results are presented to verify the theoretical results in Section V, which is followed by the concluding remarks in Section VI.

\emph{Notations.} The vectors and  matrices are respectively denoted by lowercase letters and uppercase letters. 
The operator $\odot$ denotes the Hadamard product, and the operator $\E{\cdot}$ denotes the expectation.
The operators $|\cdot|$ and $\|\cdot\|$ respectively denote the absolute value and the $\ell_2$-norm. 
The terms $\mathbb{C}$ and $\mathbb{R}$ are the set of complex values and the set of real values, respectively.
	
\section{Preliminaries}
\subsection{One-shot Distributed PCA}\label{sec:pca}
%The dimension of datasets keeps increasing in recent years. 
%A surging dimension of datasets results in the ever-increasing model size of machine learning tasks. 
%Therefore, a feature extraction procedure is introduced to reduce the dimension of datasets for supervised learning tasks such that the model size and training complexity can be reduced. 
As one of the dimension reduction algorithms, PCA is widely used due to its simplicity in implementation and the preservation of the most characteristics of the original dataset \cite{jolliffe2005principal}.
Note that the PCA requires to perform singular-value decomposition (SVD) to the original dataset $A \in \mathbb{R}^{d_0 \times M}$ with $d_0$ and $M$ as the dimension and number of samples, respectively.
However, the SVD operation cannot be implemented in WFL systems.
The reasons are as follows: 1) the original dataset $A$ is split across multiple workers; and 2) the server is not allowed to collect data samples from workers in the WFL systems. 
Although an iterative distributed PCA is proposed in \cite{9749966}, a one-shot distributed PCA \cite{liang2014improved} is preferred in the WFL systems to save the scarce radio resource. 

%Denote an $M$-sample dataset by $A \in \mathbb{R}^{d_0 \times M}$, where $d_0$ is the dimension of each sample. 
%The PCA starts with performing the singular-value decomposition (SVD) to the original dataset $A$ and selects the top $d_1$ left-singular vectors to span a $d_1$-dimensional subspace with $d_1 < d_0$.
%Then, the PCA retrieves the low-dimensional dataset $\hat A \in \mathbb{R}^{d_1 \times M}$ by projecting the original dataset $A$ onto the $d_1$-dimensional subspace.  	

To facilitate the investigation of a WFL system with high-dimensional datasets, we start with reviewing the one-shot distributed PCA. We refer readers to sources \cite{liang2014improved, feldman2020turning} for more details.  
Suppose that each worker $n$ has $M$ samples as $A_n \in \mathbb{R}^{d_0 \times M}$. 
The original dataset $A$ is thus denoted by $A = [A_1, A_2, \ldots, A_N] \in \mathbb{R}^{d_0 \times M N}$.
We summarize the \mbox{one-shot} distributed PCA in Algorithm \ref{alg:distpca}. 
	
\begin{algorithm}[htb]  
\centering
\caption{One-shot Distributed PCA Algorithm}\label{alg:distpca}
\begin{algorithmic}[1]
	\State Each worker $n$ performs the  SVD to local dataset $A_n = U_n\Lambda_n V_n^{T}$
	\State Each worker $n$ uploads the first $\hat d_0$ columns of $U_n \Lambda_n$ to the server
	\State The server performs the SVD to the collected matrix $[U_1 \Lambda_1, U_2 \Lambda_2, \ldots, U_N \Lambda_N] = U\Lambda V^{T}$
	\State The server obtains the first $\hat d_0$ columns of $U$ as $\hat U = U[1:\hat d_0]$
	\State The server broadcasts the matrix $\hat U$ to $N$ workers
	\State Each worker $n$ performs the dimension reduction of local dataset as $\hat A_n = \hat U^{T} A_n$
	\end{algorithmic}
\end{algorithm}

When $\hat d_0$ is carefully selected, each low-dimensional local dataset $\hat A_n \in \mathbb{R}^{\hat d_0 \times M}$ can preserve the most important information for training \cite[Theorem 9]{liang2014improved}. 

\begin{figure}[tb]
	\centering
	\includegraphics[width=2.3 in]{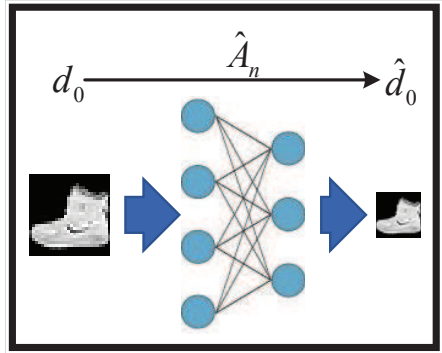}
	\caption{Mapping a $d_0$-dimensional figure to a $\hat d_0$-dimensional figure based on the one-shot distributed PCA.}
	\label{fg:dist_pca}
\end{figure}

We perform a budget calculation to illustrate the benefits of the one-shot distributed PCA as in Fig. \ref{fg:dist_pca}. 
We consider to train a three-layer perceptron to classify the FMNIST dataset \cite{xiao2017fashion}. 
Suppose that the input, hidden, and output layers respectively have $d$, $64$, and $10$ neurons with $d$ denoting the dimension of the data sample. 
The numbers of model parameters between the input layer and the hidden layer are, respectively, $64(d_0 + 1)$ and $64 (\hat d_0 + 1)$ before and after the \mbox{one-shot} distributed PCA. 
When both $\hat d_0$ and $d_0$ are sufficiently large with $\hat d_0 \le d_0$, the dimension of exchanged information is reduced by approximately $1 - \nicefrac{\hat d_0}{d_0}$ per frame.
Note that $d_0$ is $784$ for the FMNIST dataset. 
When we set $\hat d_0 = 500$, the communication saving is $36.22\%$ per frame. 
We will illustrate the impacts of the dimension reduction on the learning performance in the numerical results.  

{\color{blue}The one-shot distributed PCA algorithm can reduce the demand for wireless channels of our proposed algorithms, and thereby increase the potential application scenarios.}

\subsection{Intelligence Quantification}\label{sectioniib}
Different generations of networking technologies establish cooperation of human beings by allowing the movement of ``something'', such as matter, energy, and information  \cite{9448012}. 
The next generation of networking technology is built upon a higher level abstraction of the current one. 
For example, the energy quantifies a matter's moving speed, and information quantifies the energy dispersal speed. 
Therefore, we can envision that a new networking technology can provide a higher abstraction of information, namely, intelligence. 
Mimicking the ways to define energy in Thermodynamics and information in Shannon's entropy, it is suggested that the difference of intelligence can be defined as $d w = \nicefrac{d S}{d R}$ where $S$ denotes the similarity of the current and the optimal orders, and $R$ is the concerned parameter  \cite{9448012}. 
When the concerned parameter $R$ is time in a supervised learning task, the difference of intelligence is specified as $\nicefrac{d w}{d t} = -\nabla f (w)$ with the gradient of empirical risk function $\nabla f(w)$ \cite{murray2019revisiting}.
Therefore, we are motivated to perform the finite-time analysis of the time-average variance of intelligence's difference to quantify the spread speed of intelligence in the investigated WFL system.

\section{System Model and Problem Description}
{\color{blue}
In this section, we present an analog WFL system where multiple workers collaboratively learn a common model under the orchestration of a server.
Since the difference of intelligence is contained in the gradient as discussed in Section \ref{sectioniib}, we consider that the $N$ workers upload the local gradients to the server for the model update over orthogonal wireless channels. 
After the global model update, the $N$ workers download the global model from the server as shown in Fig. \ref{fg:senario}. 

\begin{figure}[tb]
	\centering
	\includegraphics[width=3.1 in]{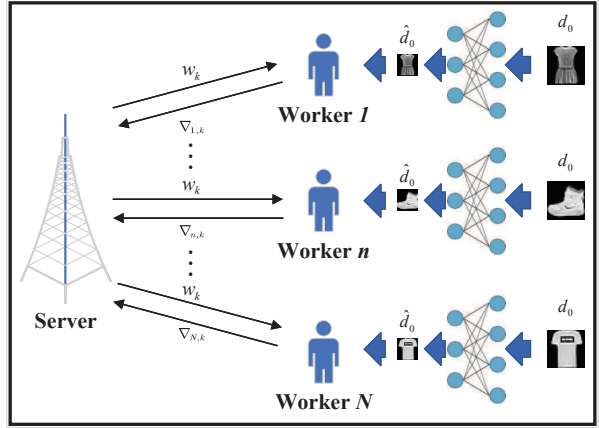}
	\caption{An illustration of WFL system with one-shot distributed PCA.}
	\label{fg:senario}
\end{figure}

\subsection{Problem Description}
We consider a supervised learning task in the analog WFL system, where the server and $N$ workers coordinately minimize the empirical risk as 
\begin{equation}\label{eqa:01}
\min_{w} f(w) :=  \frac{1}{N}\sum_{n=1}^N f_n(w)
\end{equation}
where $f_n(w) = \frac{1}{M}\sum_{m=(n-1)M}^{nM-1}\ell(w;\hat a_{m,n})$ is the local loss of  worker $n$ with respect to the model $w \in \mathbb{R}^{d_1}$ and the sample $\hat a_{m,n} \in \mathbb{R}^{\hat d_0}$.
More specifically, each $\hat a_{m,n}$ is the $m$th column of dataset $\hat A_n$. 

\subsection{Signal Model for Aggregated Gradient}
In this work, we consider the non-line of sight Rayleigh fading uplink channels and the reliable downlink channel.
Thus, each element of channel coefficient vector $h_{n,k}:= [h_{n,k}[i]]_{i=1}^{d_1} \in \mathbb{C}^{d_1}$ follows an i.i.d. circularly symmetric complex Gaussian distribution ${\cal CN}(0, \delta_{n,k}^{-\alpha} I)$ with the propagation distance $\delta_{n,k}$ and the pathloss exponent $\alpha$. 
Here, the value of $d_1$ denotes the number of channel usages and is equal to the dimension of the local gradient. 

During the downlink period of frame $k$, the server broadcasts the model $w_k \in \mathbb{R}^{d_1}$ to all workers.
In the uplink period per frame $k$, we allow each worker $n$ to upload the $n$th local normalized gradient $\nicefrac{ y_{n,k} }{\| y_{n,k} \|}$ with $y_{n,k} :=  \nabla f_n(w_k) \in \mathbb{R}^{d_1}$ to the server as
\begin{equation}\label{eqa:02}
	\nabla_{n,k} = h_{n,k} \odot p_{n,k} \odot \frac{ y_{n,k} }{\| y_{n,k} \|} + z_{n,k}, \forall n
\end{equation}
where $p_{n,k} := [p_{n,k}[i]]_{i=1}^{d_1} \in \mathbb{C}^{d_1}$ is the power control vector of worker $n$ per frame $k$ with the $i$th element as $p_{n,k}[i]$, the operator $\odot$ is the Hadamard product of two matrices, and $z_{n,k}$ is the real part of AWGN with mean zero and covariance matrix $\!\sigma^{2} I \!$.

Following the uplink transmission in \eqref{eqa:02}, the elements of the $n$th local normalized gradient are transmitted over orthogonal channel usages per frame $k$. 
Given the $n$th local gradient $y_{n,k}$, the transmit power of  worker $n$ over all channels is limited by 
\begin{equation}\label{eqa:03}
\Big\| p_{n,k} \odot \frac{ y_{n,k} }{\| y_{n,k} \|} \Big\|^2 = p_0, \forall n
\end{equation}
where $p_0$ is the power budget of each radio front-end. 

When the $i$th channel usage of worker $n$ experiences deep fading (i.e., $|h_{n,k}[i]| < h_0$) with $h_0$ as the predetermined truncation threshold, the $i$th channel usage cannot be used for information transmission by worker $n$. 
We define the scheduling policy by setting each element in ${\one}_{n, k} \in \mathbb{R}^{d_1}$ to one when the corresponding channel usage can be used by worker $n$ per frame $k$ (i.e., $|h_{n,k}[i]| \ge h_0$), and to zero otherwise. 
The expectation of each element ${\one}_{n, k}[i]$ is obtained as $\E{ \one_{n,k}[i]  } = \mbox{Pr}[ |h_{n,k}[i]| \ge h_0  ] = \exp(-\delta_{n,k}^{\alpha} h_0^2)$, which depends on the propagation distance of worker $n$ and the truncation threshold $h_0$.
Since different channel usages experience an i.i.d. fadings, the worker $n$ needs to avoid scaling different elements of the $n$th local normalized gradient by different channel power gains. 
Therefore, the workers perform power control such that each element in the $n$th received local normalized gradient is expected to be aligned with the same positive power control factor $\rho_{n,k}$. 
Different from \cite{9272666, 8870236, 9252950}, we perform channel alignment by considering the channel scheduling probability as 
\begin{equation}\label{eqa:04}
	p_{n,k}[i] \!=\! \left\{
	\begin{matrix}
		\rho_{n, k} c_{n,k} \frac{h^*_{n,k}[i]}{|h_{n,k}[i]|^2},\!\! &|h_{n,k}[i]| \ge h_0\\
		0,\!\! &|h_{n,k}[i]| < h_0
	\end{matrix}
	\right.
\end{equation}
where $c_{n,k}$ is the inverse of channel scheduling probability with $c_{n,k} := 1/\E{ \one_{n,k}[i]  } =  \exp(\delta_{n,k}^{\alpha} h_0^2) $.

Substituting channel alignment policy \eqref{eqa:04} into \eqref{eqa:02}, the $n$th intermittent gradient is obtained as 
\begin{equation}\label{eqa:07}
	\nabla_{n,k} = \rho_{n, k} c_{n,k} \one_{n, k} \odot \frac{y_{n,k}}{\|y_{n,k}\|} + z_{n,k}, \forall n.
\end{equation}

Based on \eqref{eqa:07}, we have $\E{ 	\nabla_{n,k} } = \rho_{n,k} \nicefrac{y_{n,k}}{\|y_{n,k}\|}$ by recalling the facts $\E{ \one_{n,k}[i]  } =  \exp(-\delta_{n,k}^{\alpha} h_0^2)$ and $c_{n,k} = 1/\E{ \one_{n,k}[i]  } =  \exp(\delta_{n,k}^{\alpha} h_0^2) $.
Note that the OAC technology only allows one single power control factor in the system, i.e., $\rho_k \equiv \rho_{n, k}$.
Confined by the near-far effect of co-channel transmission, the received strength of local gradients is controlled by the farthest worker in the analog WFL systems \cite{8870236, 9272666}.
Therefore, the received strength for all workers can be extremely low when the propagation distance of the farthest worker is very large. 
In the investigated analog WFL system, each worker $n$ uploads intermittent gradient $\nabla_{n,k}$ over orthogonal channel usages.
Compared with the OAC technology, the investigated analog WFL system can use different power control factors $[\rho_{n, k}]_{n=1}^N$ to mitigate the the near-far effect of co-channel transmissions.
We refer the removal of channel alignment factors as gradient alignment. 
After performing the gradient alignment by $\nicefrac{\| y_{n,k} \| }{ \rho_{n,k}}$, the $n$th received gradient is
\begin{equation}\label{eqa:08}
	\hat \nabla_{n,k} = c_{n,k} \one_{n, k} \odot y_{n,k}  + \frac{ z_{n,k} }{  \rho_{n,k} } \| y_{n,k} \|, \forall n.
\end{equation}

Denote the accumulated noise per frame $k$  by $z_k = \frac{1}{N}\sum_{n=1}^N \nicefrac{z_{n,k} \| y_{n,k} \| }{ \rho_{n,k} } $ with $\E{ z_k } = 0$.
Based on \eqref{eqa:08}, we obtain the aggregated gradient as
\begin{equation}\label{eqa:09}
	\frac{1}{N}\sum_{n=1}^N \hat \nabla_{n,k} = \frac{1}{N}\sum_{n=1}^N c_{n,k}\one_{n, k} \odot y_{n,k} + z_k.
\end{equation}

Different from i.i.d. assumption on the propagation distance of workers \cite{8870236, 9562537, 9014530, 9272666}, the received gradients in \eqref{eqa:09} show that our investigated analog WFL system factors in the propagation distances of workers. }

\section{Algorithm Design and Convergence Analysis}
In this section, based on the following conditions on the local loss functions $f_n$, $n =1, 2, \ldots, N$, we develop the \mbox{PCA-WFL} algorithm and its accelerated version.
Subsequently, we will analyze the convergence of our proposed algorithms.

\begin{condition}[Bounded Variance of Gradient]\label{cd:01}
	Each local loss function is differentiable and has bounded gradient variance as $\frac{1}{N}\sum_{n=1}^N\|y_{n,k} - \nabla f(w_k)\|^2 \le G$. 
\end{condition}

\begin{condition}[Smoothness {\cite[eq. (1.2.11)]{nesterov2018lectures}}]\label{cd:02}
	Each local loss function has $L$-Lipschitz continuous gradient as 
	\begin{equation}\label{eqa:12}
		f_n( u ) \le f_n(w) + \inp{\nabla f_n(w), u - w} + \frac{L}{2}\| u - w \|^2
	\end{equation}
	where $u  \in \mathbb{R}^{d_1}$ and $w \in \mathbb{R}^{d_1}$. 
\end{condition}

Condition \ref{cd:01} quantifies the deviation of local gradients from the global gradient \cite{9272666}. 
Besides, Condition \ref{cd:01} is milder than the assumption on the bounded gradient in \cite{9252950, 9042352}.
Condition \ref{cd:02} requires the loss functions to be smooth and can be easily satisfied for some artificial neural networks \cite{9252950, latorre2020lipschitz, 9042352}.

Before starting the algorithm development, we introduce Lemma \eqref{le:02} to characterize the first-order and the second-order statistical properties of the aggregated gradient estimator \eqref{eqa:09}. 
 
\begin{lemma}\label{le:02}
When Condition \ref{cd:01} is satisfied, the aggregated gradient \eqref{eqa:09} is an unbiased estimator of global gradient as 
\begin{equation}\label{eqa:05}
\EE{ \frac{1}{N}\sum_{n=1}^N \hat \nabla_{n,k} } = \frac{1}{N}\sum_{n=1}^N y_{n,k}.
\end{equation}

The upper bound of the aggregated gradient per frame $k$ is 
\begin{equation}\label{eqa:10}
\!\!\EE{ \| \frac{1}{N}\!\!\sum_{n=1}^N \! \hat\nabla_{n,k} \|^2 } 
\!\le\!  2\E{ \|\nabla f(w_k)\|^2 }  + \frac{c_1 \!+\! c_2 d_1\sigma^2 \!+\! 4G}{ 2N}\!\!
\end{equation}
where $c_1 := \frac{1}{N} \sum_{n=1}^N [\exp(\delta_{n,k}^{\alpha} h_0^2) - 1]$ and  $c_2 := \frac{1}{Np_0} \sum_{n=1}^N \delta_{n,k}^{\alpha} \exp(2\delta_{n,k}^{\alpha} h_0^2) E_1( \delta_{n,k}^{\alpha}h_0^2 )$. 
\end{lemma}
\begin{IEEEproof}
See Appendix \ref{apdx:02}.
\end{IEEEproof}

Based on \eqref{eqa:05}, Lemma \ref{le:02} shows that the aggregated gradient \eqref{eqa:09} is an unbiased estimator of the global gradient when channel alignment policy \eqref{eqa:04} includes the channel scheduling policy. 
Based on \eqref{eqa:10}, Lemma \ref{le:02} also shows that the variance of the aggregated gradient is positively correlated to the propagation distances of workers $[\delta_{n,k}]_{n=1}^N$ and the truncation threshold $h_0$.

\subsection{PCA-WFL Algorithm}
Using the aggregated gradient \eqref{eqa:09} and the one-shot distributed PCA, we present the PCA-WFL algorithm as 
\begin{equation}\label{eqa:11}
	w_{k+1} = w_k - \eta \frac{1}{N}\sum_{n=1}^N \hat \nabla_{n,k}
\end{equation}
where $\eta$ is a stepsize.

We perform a finite-time analysis to obtain the convergence rate of the PCA-WFL algorithm in the following theorem, whose proof relies on Lemma \ref{le:02}. 

\begin{theorem}\label{th:vanilla}
	Assume that Conditions \ref{cd:01} and \ref{cd:02} are satisfied. 
	When the stepsize $\eta$ satisfies $\eta \le \nicefrac{3}{4 L}$, the convergence rate of the PCA-WFL algorithm is 
	\begin{equation}\label{eqa:13}
	\frac{1}{K}  \sum_{k=0}^{K-1} \E{ \| \nabla f(w_k) \|^2 } 
	= {\cal O}(\frac{1}{\eta K}) + {\cal O}(\frac{\eta}{N})
	\end{equation}
	where ${\cal O}(\frac{1}{\eta K})$ and ${\cal O}(\frac{\eta}{N})$ are respectively defined as 
	\begin{equation}\label{eqa:14}
	{\cal O}(\frac{1}{\eta K}) := \frac{ 4 }{\eta K} \E{ f(x_{0}) -\! f^* }
	\end{equation}
	and
	\begin{equation}\label{eqa:15}
	{\cal O}(\frac{\eta}{N}) :=  (c_1 \!+\! c_2 d_1 p_0^{-1} \sigma^2 \!+\! G) \frac{\eta L}{N}. 
	\end{equation}
\end{theorem}
\begin{IEEEproof}
	See Appendix \ref{apdx:03}.
\end{IEEEproof}

Theorem \ref{th:vanilla} quantifies the variance of empirical risk's gradient over the total number of frames $K$, which reflects the spread of intelligence over different workers in finite-time horizon $K$ for the PCA-WFL algorithm.  
The right-hand side (RHS) of \eqref{eqa:13} shows that the PCA-WFL algorithm converges at a rate ${\cal O}(\nicefrac{1}{\eta K})$ to an ${\cal O}(\nicefrac{\eta}{N})$-neighbor (i.e., error-floor term) of the local optimal model for smooth non-convex empirical risk.
Moreover, the size of the neighborhood is controlled by the stepsize $\eta$. 
Theorem \ref{th:vanilla} will serve as a benchmark to demonstrate the performance improvement of our proposed PCA-WFL algorithm.

%\begin{figure*}[htb]
%	\begin{equation}\label{eqa:16f}
	%		\eta\beta \frac{1}{N}\sum_{n=1}^N \hat \nabla_{n,k} 
	%		= w_{k} - w_{k+1} - \eta \frac{1}{N} \sum_{n=1}^N \hat \nabla_{n,k} 
	%		- \beta\sq{ w_{k-1} - w_k - \eta \frac{1}{N}\sum_{n=1}^N  \hat \nabla_{n,k-1} }
	%	\end{equation}
%\hrulefill
%\end{figure*}

\subsection{Accelerating the PCA-WFL Algorithm}
In this subsection, we leverage the prestigious Nesterov's momentum to accelerate the PCA-WFL algorithm.
We start by giving an intuition on the acceleration capability of the Nesterov's moment based on Section \ref{sectioniib}. 

Recalling \cite[eqs. (22)--(23)]{yuan2016influence}, the standard Nesterov's momentum recursion is
\begin{subequations}\label{eqa:16d}
\begin{align}
	u_{k} &= w_k -  \eta\frac{1}{N}\sum_{n=1}^N \hat \nabla_{n,k} \label{eqa:16db}\\
	w_{k+1} &= (1+\beta) u_k - \beta u_{k-1} \label{eqa:16da}
\end{align}
\end{subequations}
where $u_k$ is a sequence of auxiliary variables. 

%{\color{blue}
%\begin{figure}[ht]
%	\centering
%\includegraphics[width=3.3 in]{intuition_nesterov.pdf}
%\caption{The intuition of standard Nesterov's recursion. The models $\hat w_{k+1}$ and $\hat  w_{k+2}$ are obtained via the vanilla gradient descent, and the models $w_{k+1}$ and $w_{k+2}$ are obtained via the standard Nesterov's momentum.}\label{fig:intuition}
%\end{figure}
%
%As shown in Fig. \ref{fig:intuition}, the standard Nesterov's recursion corrects the direction of vanilla gradient  via \eqref{eqa:16da} after a one-step gradient descent in \eqref{eqa:16db}. Therefore, the standard Nesterov's recursion accelerates the convergence by allowing the effective gradient points more accurate to the optimal point compared with the vanilla gradient descent. }

However, the standard Nesterov's recursion in \eqref{eqa:16d} cannot intuitively illustrate the speedup capability of the momentum term. 
Therefore, we provide an equivalent recursion of \eqref{eqa:16d} as follows. 
Substituting \eqref{eqa:16db} into \eqref{eqa:16da} and performing several algebraic manipulations, we obtain
\begin{equation}\label{eqa:16f}
\eta\beta \frac{1}{N}\sum_{n=1}^N \hat \nabla_{n,k} 
= w_{k} - w_{k+1} - \eta \frac{1}{N} \sum_{n=1}^N \hat \nabla_{n,k} 
- \beta\sq{ w_{k-1} - w_k - \eta \frac{1}{N}\sum_{n=1}^N  \hat \nabla_{n,k-1} }. 
\end{equation}

Setting 
\begin{equation}\label{eqa:16g}
\eta\beta u_{k+1} := w_{k} - w_{k+1} - \eta \frac{1}{N}\sum_{n=1}^N  \hat \nabla_{n,k}
\end{equation}
and
\begin{equation}\label{eqa:16h}
v_k := \beta u_{k+1} + \frac{1}{N}\sum_{n=1}^N  \hat \nabla_{n,k}.
\end{equation}

Based on \eqref{eqa:16f}--\eqref{eqa:16h}, we obtain $w_{k+1} = w_k - \eta v_k$ and $u_{k+1} = \beta u_k +  \frac{1}{N}\sum_{n=1}^N \hat \nabla_{n,k}$. 
Therefore, we recast the recursion \eqref{eqa:16d} as
\begin{subequations}\label{eqa:16}
\begin{align}
u_{k+1} &= \beta u_{k} + \frac{1}{N}\sum_{n=1}^N \hat \nabla_{n,k} \label{eqa:16a}\\
v_{k} &= \beta u_{k+1} + \frac{1}{N}\sum_{n=1}^N \hat \nabla_{n,k} \label{eqa:16b} \\
w_{k+1} &= w_k - \eta v_{k} \label{eqa:16c}
\end{align}
\end{subequations}
where $\beta$ is a momentum factor.

Applying the recursion in \eqref{eqa:16a} with  $u_0 = 0$, we obtain
\begin{equation}\label{apdx04:04}
	u_k = \sum_{t = 0}^{k-1} \beta^{k-1-t} \frac{1}{N}\sum_{n=1}^N \hat\nabla_{n,t}.
\end{equation}

Substituting \eqref{eqa:16b} and \eqref{apdx04:04} into \eqref{eqa:16c}, we obtain 
\begin{equation}\label{eqa:16i}
\begin{split}
&\frac{1}{\eta}(w_{k} - w_{k+1}) \\
&=  (1+\beta) \frac{1}{N}\sum_{n=1}^N \hat\nabla_{n,k} +  \sum_{t = 0}^{k-1} \beta^{k+1-t} \frac{1}{N}\sum_{n=1}^N \hat\nabla_{n,t}.
\end{split}
\end{equation}

We observe that \eqref{eqa:16i} updates model $w_k$ via a momentum-based aggregated gradient and the discounted historical aggregated gradients while the PCA-WFL algorithm only uses the received gradient. 
Section \ref{sectioniib} indicates that the spread of intelligence is contained in the gradient of empirical risk, which is the aggregated gradient in the investigated analog WFL system. 
Therefore, the Nesterov's momentum can speed up the PCA-WFL algorithm by exploiting more intelligence from the historical aggregated gradients. 

Combining \eqref{eqa:16} with the one-shot distributed PCA, we summarize the procedures of our proposed \mbox{PCA-AWFL} algorithm in Algorithm \ref{alg:01}.
Different from \cite{9042352}, exchanging the normalized gradients reduces the steps for mean value removal for the received gradients.

\begin{algorithm}[htb]  
\centering
\caption{PCA-AWFL Algorithm}\label{alg:01}
\begin{algorithmic}[1]
\State \textbf{Initialization:} Momentum factor $\beta$, stepsize $\eta$, and initial values $u_0  = 0$
\State The server and  workers cooperatively perform the Algorithm \ref{alg:distpca} to reduce the dimension of local datasets
\For{$k = 1, \ldots, \infty$}
\State The server broadcasts the model vector $w_k$ to all  workers
\State Each worker $n$ estimates the channel coefficient vector $h_{n,k}$ and the inverse of channel scheduling probability $c_{n,k}$
\State Each worker $n$ sets the channel alignment factor as 
\begin{equation}
\frac{1}{\rho^{2}_{n,k}} =  \frac{c_{n,k}^2}{p_0 } \sum_{i=1}^{d_1}\frac{ |\one_{n,k}[i]y_{n,k}[i]|^2 }{|h_{n,k}[i]|^2\|y_{n,k}\|^2}
\end{equation}
\State Each worker $n$ performs the channel alignment in \eqref{eqa:04}
\State Each worker $n$ uploads to the server the local normalized gradient $\nicefrac{ y_{n,k} }{\| y_{n,k} \|}$
\State The server performs gradient alignment by using $[\nicefrac{\| y_{n,k}\| }{\rho_{n,k}} ]_{n=1}^N$ and updates model via \eqref{eqa:16}
\EndFor
\end{algorithmic}
\end{algorithm}

\begin{remark}[Link Budget Analysis]
Each worker $n$ needs to upload the local normalized gradient $\nicefrac{y_{n,k}}{\| y_{n,k} \|}$ and the gradient alignment factor $\nicefrac{\| y_{n,k} \|}{\rho_{n,k}}$ to the server. 
Therefore, the channel useage per worker $n$ is $d_1 + 1$, and the total channel useage for each global gradient aggregation is $N(d_1 + 1)$.
\end{remark}

Leveraging Lemma \ref{le:02}, we establish the convergence rate of our proposed PCA-AWFL algorithm. 

\begin{theorem}\label{th:01}
Assume that the Conditions \ref{cd:01} and \ref{cd:02} are satisfied, and $u_0 = 0$. 
When the stepsize $\eta$ and momentum factor $\beta$ satisfy $\eta \le \nicefrac{3(1-\beta)^2}{ 2(2 + \beta^3) L}$, the PCA-AWFL algorithm converges at a rate of
\begin{equation}\label{eqa:17}
\frac{1}{K}  \sum_{k=0}^{K-1} \E{ \| \nabla f(w_k) \|^2 } 
= {\cal O}(\frac{1-\beta}{\eta K}) + {\cal O}(\frac{\eta}{N(1-\beta)^2})
\end{equation}
where $ {\cal O}(\frac{1-\beta}{\eta K}) $ and ${\cal O}(\frac{\eta}{N(1-\beta)^2})$ are respectively defined as
\begin{equation}\label{eqa:18}
{\cal O}(\frac{1-\beta}{\eta K}) :=  \frac{ 4(1 \!-\! \beta) }{\eta K} \E{ f(x_{0}) \!-\! f^* } 
\end{equation}
and
\begin{equation}\label{eqa:19}
{\cal O}(\frac{\eta}{N(1-\beta)^2}) := (c_1 \!+\! c_2 d_1 p_0^{-1} \sigma^2 \!+\! G ) \frac{ \eta L }{ N(1-\beta)^2 }.
\end{equation}
\end{theorem}
\begin{IEEEproof}
See Appendix \ref{apdx:04}.
\end{IEEEproof}

{\color{blue}
Based on \eqref{eqa:17}, we observe that our proposed PCA-AWFL algorithm converges to an ${\cal O}(\nicefrac{\eta}{N(1-\beta)^2})$-neighbor (i.e., the error-floor term) of local optima at a rate ${\cal O}(\nicefrac{(1-\beta)}{\eta K})$ for smooth non-convex empirical risk.
Based on \eqref{eqa:19}, we have three observations: 1) the error-floor term decreases with the number of workers $N$; 2) the error-floor term shrinks with stepsize $\eta$ at the expense of slow convergence; and 3) the error-floor term increases  with the distances of workers to the server. 
Compared with Theorem \ref{th:vanilla}, we observe from Theorem \ref{th:01} that our proposed PCA-AWFL algorithm converges faster than the PCA-WFL algorithm by a factor of $1-\beta$ at the expense of an error-floor term ${\cal O}(\nicefrac{\eta}{N(1-\beta)^2}) $-neighbor. 
When the momentum term $\beta$ reduces to zero, the PCA-AWFL reduces to PCA-WFL.

When the value $K$ is sufficiently large, we can set $\eta = \sqrt{\nicefrac{N}{K}}$ such that $\eta \le  \nicefrac{3(1-\beta)^2}{ 2(2 + \beta^3) L}$.
Based on the RHS of \eqref{eqa:17}, we can set $\eta = \sqrt{\nicefrac{N}{K}}$ such that 
\begin{equation}\label{eqa:20}
	\frac{1}{K}  \sum_{k=0}^{K-1} \E{ \| \nabla f(w_k) \|^2 } = {\cal O}(\frac{1}{\sqrt{NK}}).
\end{equation}

Eq. \eqref{eqa:20} shows that our proposed PCA-AWFL achieves a linear speedup with respect to the number of workers. 

Theorems \ref{th:vanilla} and \ref{th:01} show that the gradient norms of the \mbox{PCA-WFL} and PCA-AWFL algorithms become stable such that no more intelligence can be dispersed among the workers.
We envision that the testing accuracies of \mbox{PCA-WFL} and \mbox{PCA-AWFL} algorithms reach the peaks when the intelligence stops spreading.}

\section{Numerical Results}
{\color{blue}
We perform the numerical experiments to show the performance comparison between our proposed algorithms and the benchmarks. 
In the numerical experiments, we use the \mbox{FMNIST} dataset \cite{xiao2017fashion}, CIFAR-10 dataset \cite{Krizhevsky2009}, and AWE dataset \cite{Zhang2021}. 
The \mbox{FMNIST} dataset contains $10$-class grayscale images of fashion products with $60,000$ training samples and $10,000$ testing samples.
The CIFAR-10 dataset contains $10$-class tiny color images with $50,000$ training samples and $10,000$ testing samples.
The AWE dataset is obtained by performing a destructive experiment for automatic washing equipment of high-speed railway vehicles with $21,047$ training samples and $2,352$ testing samples.
The samples of the AWE dataset belong to six categories. 
The training samples are homogeneously distributed among the workers, and all the testing samples are placed on the server.

\subsubsection{Benchmark Descriptions} Three benchmarks are considered: the PCA based error-free batch gradient descent (PCA-EF), PCA based error-free Adam (PCA-EF-Adam), and PCA Adam (PCA-Adam) algorithms. 
More specifically, the \mbox{PCA-EF} algorithm allows the workers to reduce the dimension of data via the one-shot distributed PCA and to upload all local gradients to the server via error-free channels. 
The PCA-EF-Adam algorithm differs from the PCA-EF by using an advanced Adam solver. 
The PCA-Adam algorithm requires the workers to upload all local gradients via the Rayleigh fading channels. 

\subsubsection{Deployment Analysis} 
For the FMNIST dataset, we consider to train a three-layer perceptron with a $64$-neuron hidden layer and a $10$-neuron output layer, respectively. 
For the CIFAR-10 dataset, we consider to train a neural network that consists of a 1D-convolutional layer, a $64$-neuron hidden layer, and a $10$-neuron output layer. 
The 1D-convolutional layer contains a $7\times 1$ filter. 
For the CIFAR-10 dataset, we preprocess the data samples via a pretrained AlexNet \cite{Krizhevsky2017} for the convenience of PCA operations. 
For the AWE dataset,  we consider to train a three-layer perceptron with a $128$-neuron hidden layer and a $6$-neuron output layer, respectively. 
The hyperbolic tangent activation function is used at the hidden layer, and the loss function is the softmax cross entropy. 
The computational complexities for neural networks can be obtained based on \cite[Appendix A.1]{Molchanov}. 
In this work, we estimate the computational complexity of neural networks for three different datasets and different input sizes via the python package \emph{thop}. 
The results are shown in Table \ref{tab:01}. 
We observe that the neural network for AWE with $1,024$-dimensional input has the highest computational complexity---$131,840$ FLOPs.  
Multiplying by the number of frames, we obtain the worst-case overall complexity is around $3.96 \times 10^9$ FLOPs. 
Note that a Raspberry Pi-4B (4GB) 64-bit has a $2.02 \times 10^9$ FLOPs per second per watt\footnote{\url{https://web.eece.maine.edu/~vweaver/group/green_machines.html}}. 
The $3.96 \times 10^9$ FLOPs requirement can be well handled by current off-the-shelf energy-limited terminals that have $2.02 \times 10^9$ FLOPs per second per watt. 
Therefore, we believe that our proposed PCA-WFL and PCA-AWFL algorithms are suitable to be deployed at the off-the-shelf edge terminals in real-life applications. 

\begin{table}[htb]\footnotesize
	\centering
	\caption{Computational Complexity of Neural Networks.}\label{tab:01}
	\begin{tabular}{|c|c|c|c|c|c|}
	\hline\hline
	\multicolumn{2}{|c|}{AWE} & \multicolumn{2}{c|}{CIFAR-10} & \multicolumn{2}{c|}{FMNIST}\\
	\hline
	$\hat d_0$ & FLOPs & $\hat d_0$ & FLOPs & $\hat d_0$ & FLOPs \\	\hline
	100	 & 13,568    &100	&7,484   & 300&	19,584 \\\hline
	300	 & 39,168    &200	&14,584  & 400&	25,984 \\\hline
	500	 & 64,768    &300	&21,684  & 500&	32,384 \\\hline
	700	 & 90,368    &400	&28,784  & 600& 38,784 \\\hline
	900	 & 115,968   &500	&35,884  & 700&	45,184 \\\hline
	1024 & 131,840   &600	&42,984  & 784&	50,560  \\\hline
	\end{tabular}%
\end{table}%

\subsubsection{Parameter Setup} For the FMNIST dataset, the analog WFL system has six workers with the propagation distances as $416.33$ meters, $435.07$ meters, $389.01$ meters, $475.76$ meters, $251.43$ meters, and $163.21$ meters.
For the CIFAR-10 and AWE datasets, the analog WFL system has twelve workers with the propagation distances as $284.43$ meters, $396.79$ meters, $407.76$ meters, $444.18$ meters, $465.94$ meters, $438.90$ meters, $206.80$ meters, $435.08$ meters, $280.54$ meters, $183.13$ meters, $460.88$ meters, and $362.27$ meters.
The values of $\hat d_0$ are respectively set as $500$ for the FMNIST dataset, $400$ for the CIFAR-10 dataset, and $700$ for the AWE dataset. 
The momentum factors $\beta$ are respectively set as $0.95$ for the FMNIST dataset, $0.85$ for the CIFAR-10 dataset, and $0.9$ for the AWE dataset
The value of $h_0$ is set as $0.001$. 
The transmit power budget $p_0$ is set as $200$ mW. 
The pathloss exponent $\alpha$ is set as $2.2$. 
The noise figure at the server is $5$ dBi, and the noise figure at the workers is $15$ dBi. 
The system bandwidth is $200$ KHz, and the power spectrum density is $10^{-17.4}$ mW/Hz.

\begin{figure}[ht]
\centering
\subfigure[FMNIST with $h_0 = 0.001$.]{\includegraphics[width=0.3\linewidth]{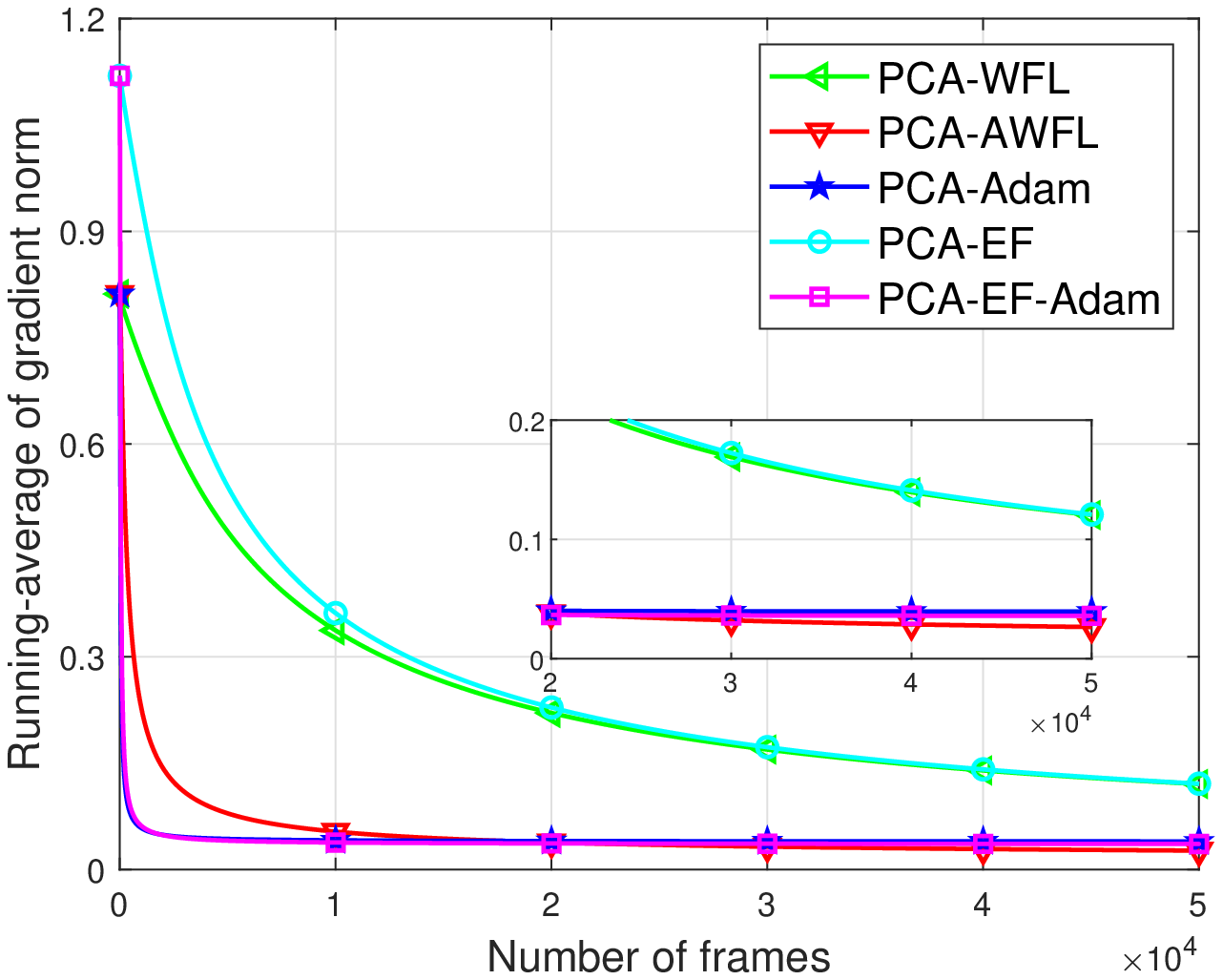}\label{fig:ra_norm0001}}\hspace{0.1cm}
\subfigure[FMNIST with $h_0 = 0.0001$.]{\includegraphics[width=0.3\linewidth]{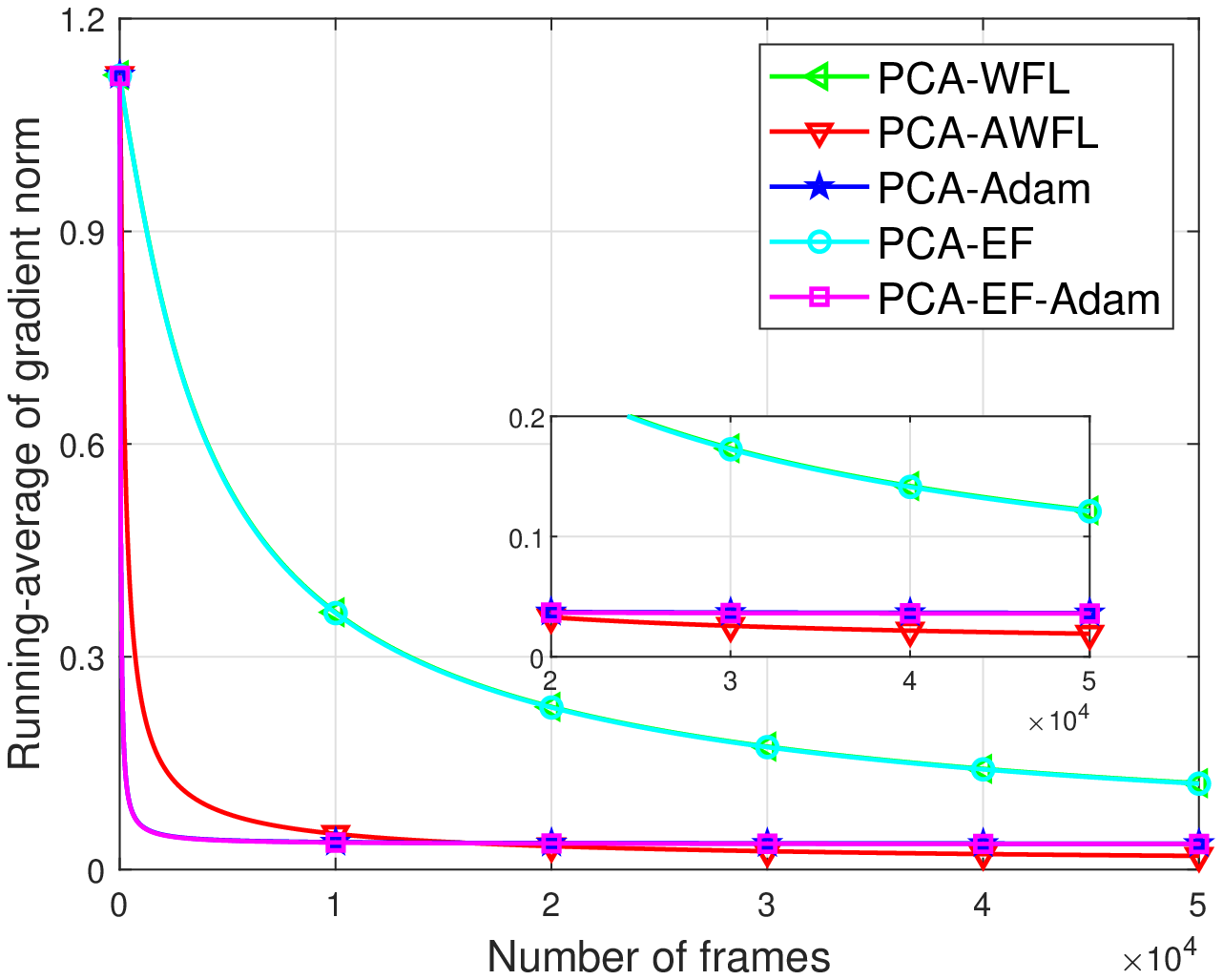}\label{fig:ra_norm00001}}
\subfigure[CIFAR-10 with $h_0 = 0.001$.]{\includegraphics[width=0.3\linewidth]{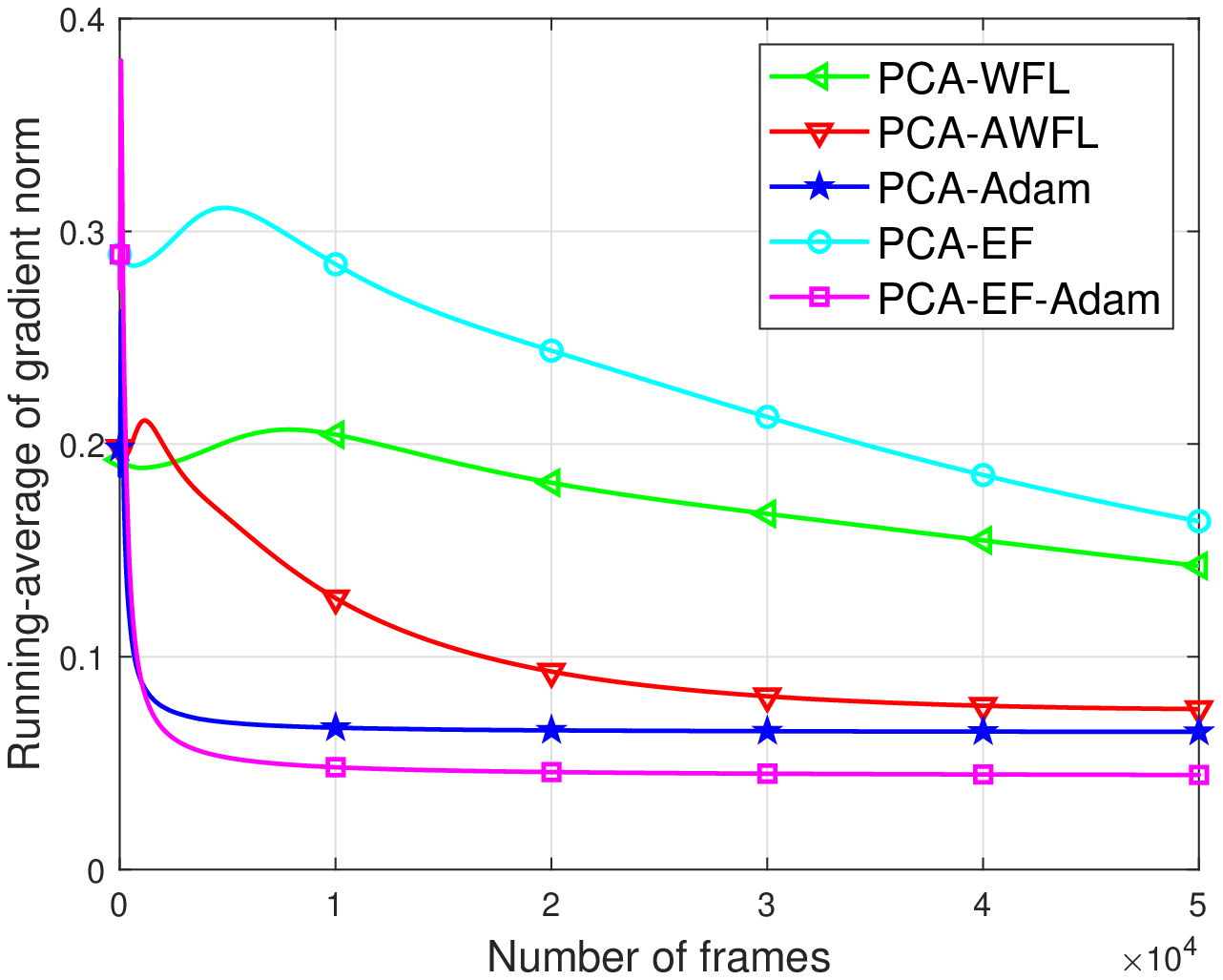}\label{fig:ra_norm0001_cifar}}\hspace{0.1cm}
\subfigure[CIFAR-10 with $h_0 = 0.0001$.]{\includegraphics[width=0.3\linewidth]{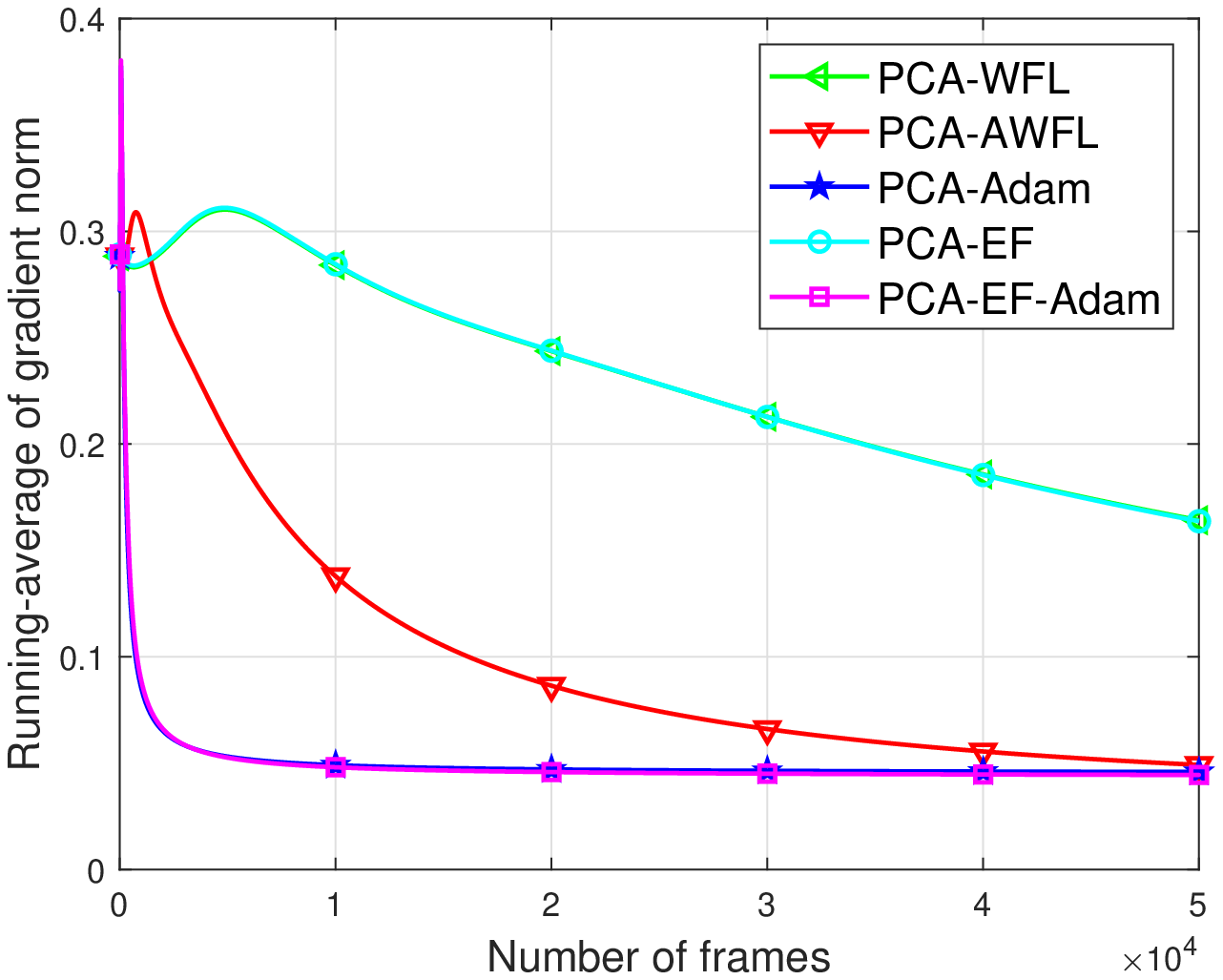}\label{fig:ra_norm00001_cifar}}
\subfigure[AWE with $h_0 = 0.001$.]{\includegraphics[width=0.3\linewidth]{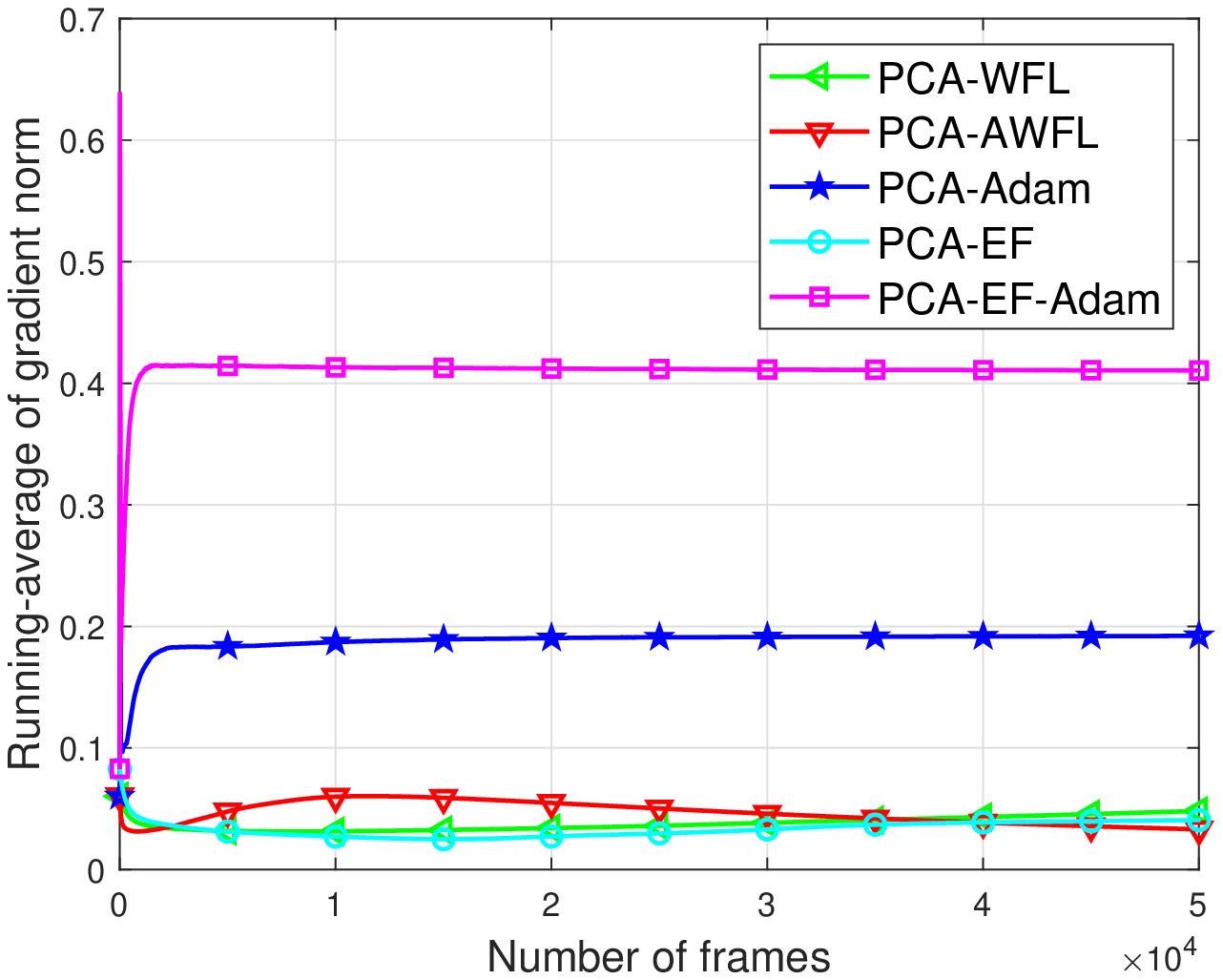}\label{fig:ra_norm0001_awe}}\hspace{0.1cm}
\subfigure[AWE with $h_0 = 0.0001$.]{\includegraphics[width=0.3\linewidth]{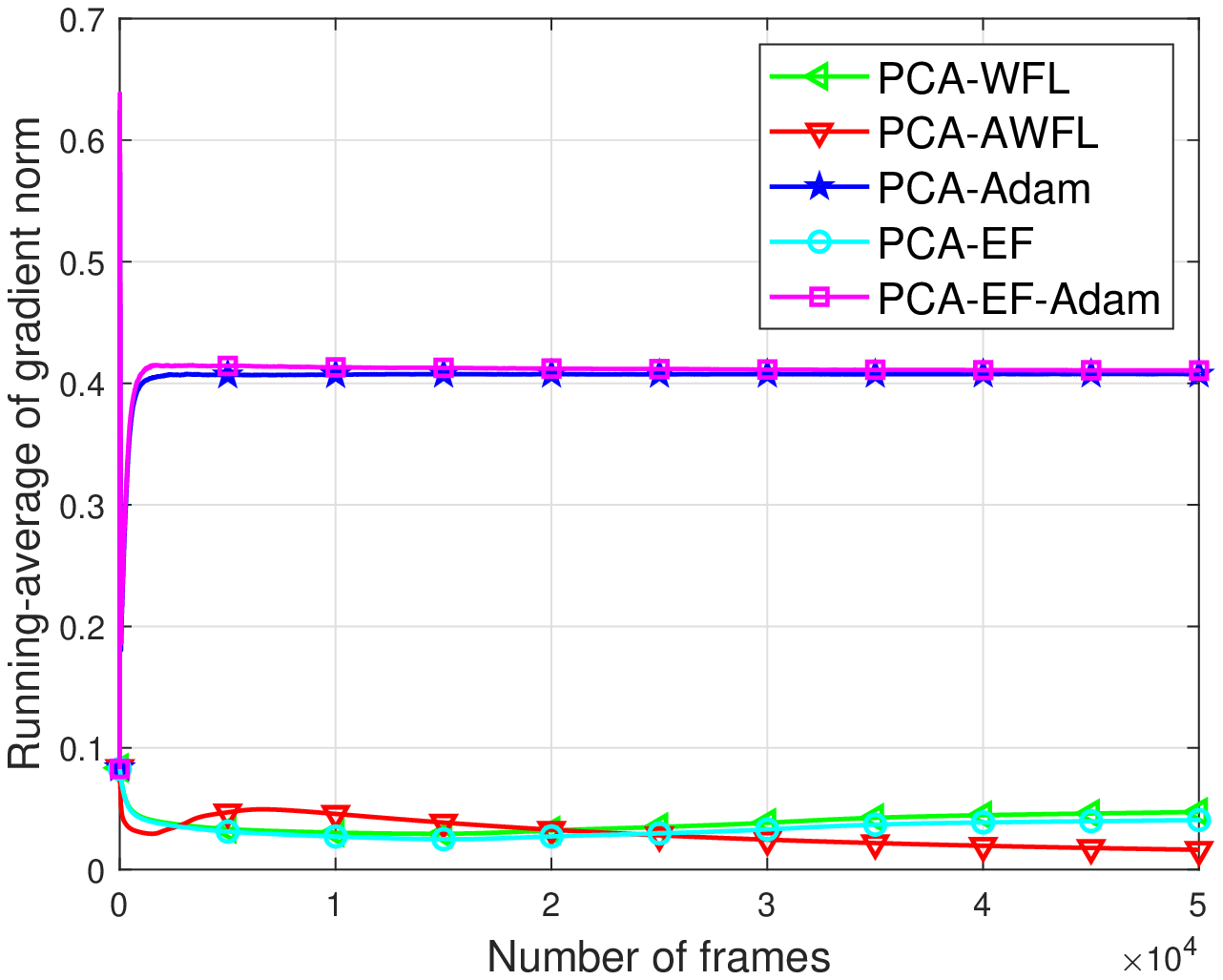}\label{fig:ra_norm00001_awe}}
\caption{The running-average of the gradient norm for the FMNIST, CIFAR-10 and AWE datasets. }\label{fig:grad_norm}
\end{figure}

Figure \ref{fig:grad_norm} shows the running-average of gradient norm after $5 \times 10^4$ frames for the PCA-WFL, PCA-AWFL, PCA-Adam, PCA-EF, and PCA-EF-Adam algorithms over FMNIST, CIFAR-10, and AWE datasets. 
Figs. \ref{fig:ra_norm0001}, \ref{fig:ra_norm0001_cifar}, and \ref{fig:ra_norm0001_awe} show the convergence of running-average of gradient norm when the truncation threshold $h_0$ is $0.001$. 
Figures \ref{fig:ra_norm00001}, \ref{fig:ra_norm00001_cifar}, and \ref{fig:ra_norm00001_awe} show the convergence of running-average of gradient norm when the truncation threshold $h_0$ is $0.0001$. 
We observe that the gradient norm of our proposed PCA-AWFL algorithm  converges after around $20,000$ frames for FMNIST, $30,000$ frames for CIFAR-10, and $30,000$ frames from AWE while the PCA-EF and PCA-WFL algorithms converge after $50,000$ frames. 
This observation verifies our theoretical results that the PCA-AWFL algorithm converges faster than the PCA-WFL algorithm in Theorem \ref{th:01}. 
Besides, we also observe that the gradient norms of PCA-Adam and PCA-EF-Adam algorithms decrease faster for the FMNIST and CIFAR-10 than the PCA-AWFL algorithm as shown in Figs \ref{fig:ra_norm0001}--\ref{fig:ra_norm00001_cifar}. 
This is because that the Adam-type algorithm benefits from the adaptive gradients \cite{Chen2019}. 
However, we observe that the gradient norms of PCA-Adam and PCA-EF-Adam algorithms converge to larger values than our proposed PCA-AWFL algorithm for the AWE dataset. 
This observation shows that myopic usage of adaptive gradients can lead to a destructive effect on the convergence behaviour.
We also illustrate the destructive effect in Figs. \ref{fig:test_acc0001_awe} and \ref{fig:test_acc00001_awe}.

\begin{figure}[ht]
\centering
\subfigure[FMNIST with $h_0 = 0.001$.]{\includegraphics[width=0.3\linewidth]{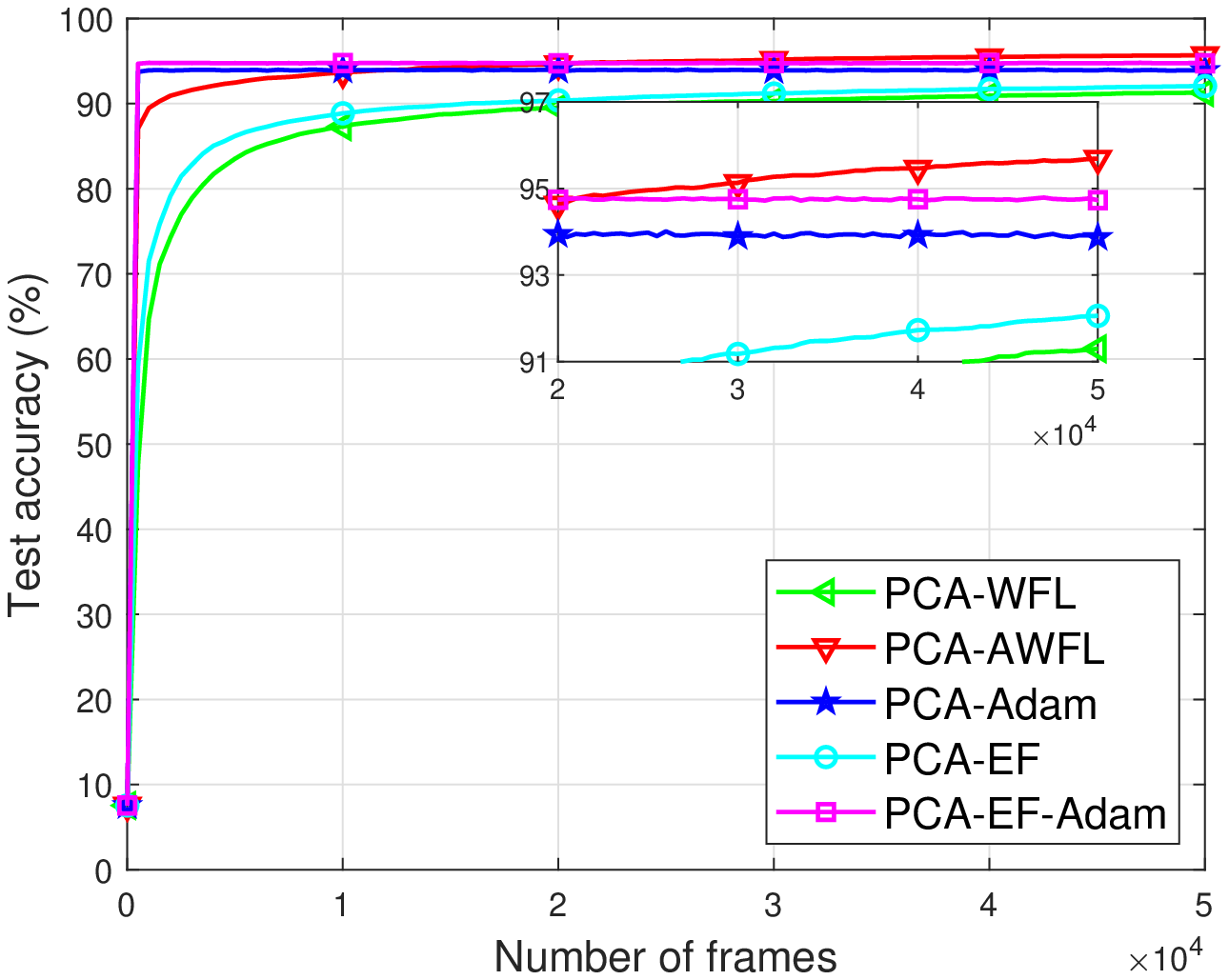}\label{fig:test_acc0001}}\hspace{0.1cm}
\subfigure[FMNIST with $h_0 = 0.0001$.]{\includegraphics[width=0.3\linewidth]{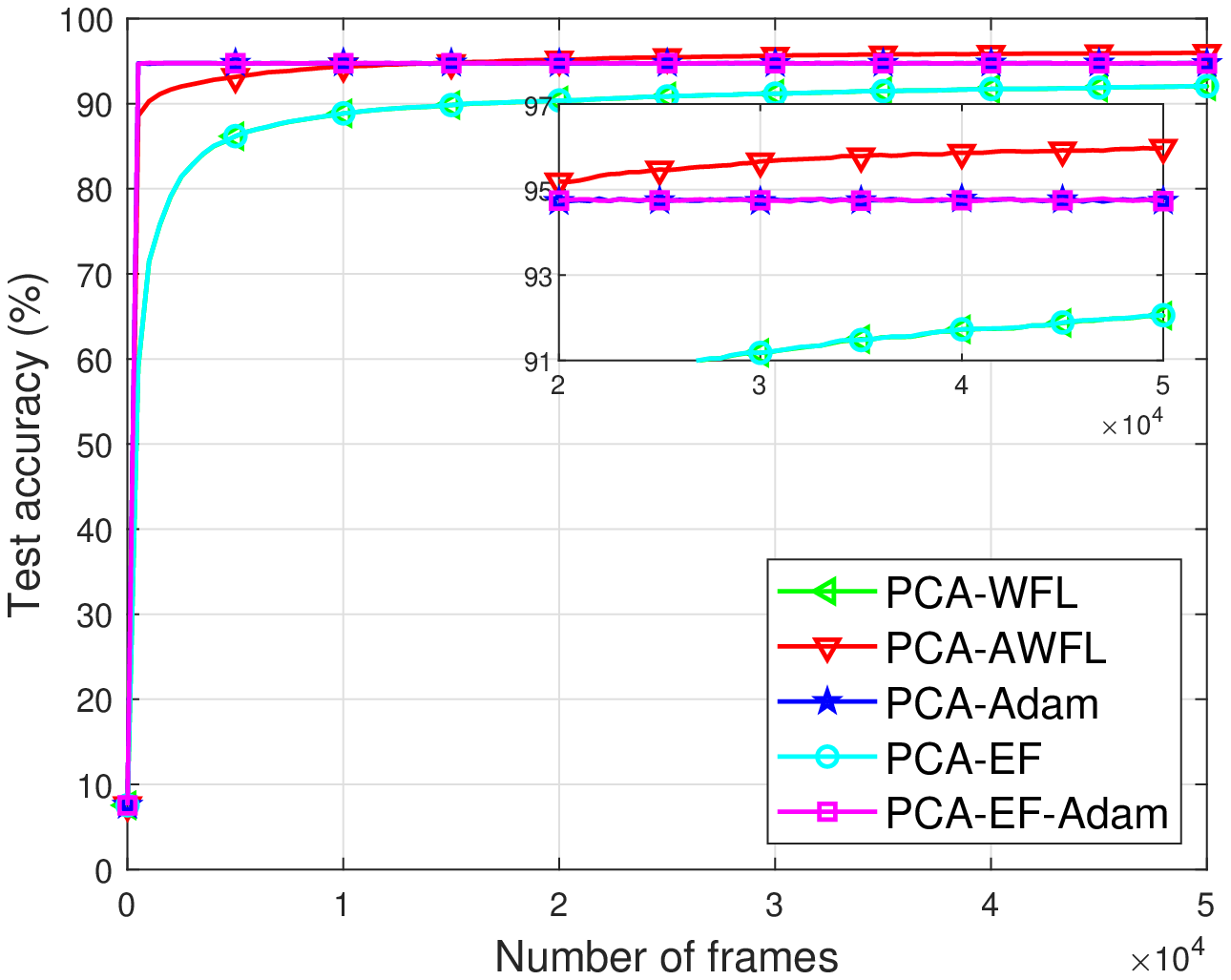}\label{fig:test_acc00001}}
\subfigure[CIFAR-10 with $h_0 = 0.001$.]{\includegraphics[width=0.3\linewidth]{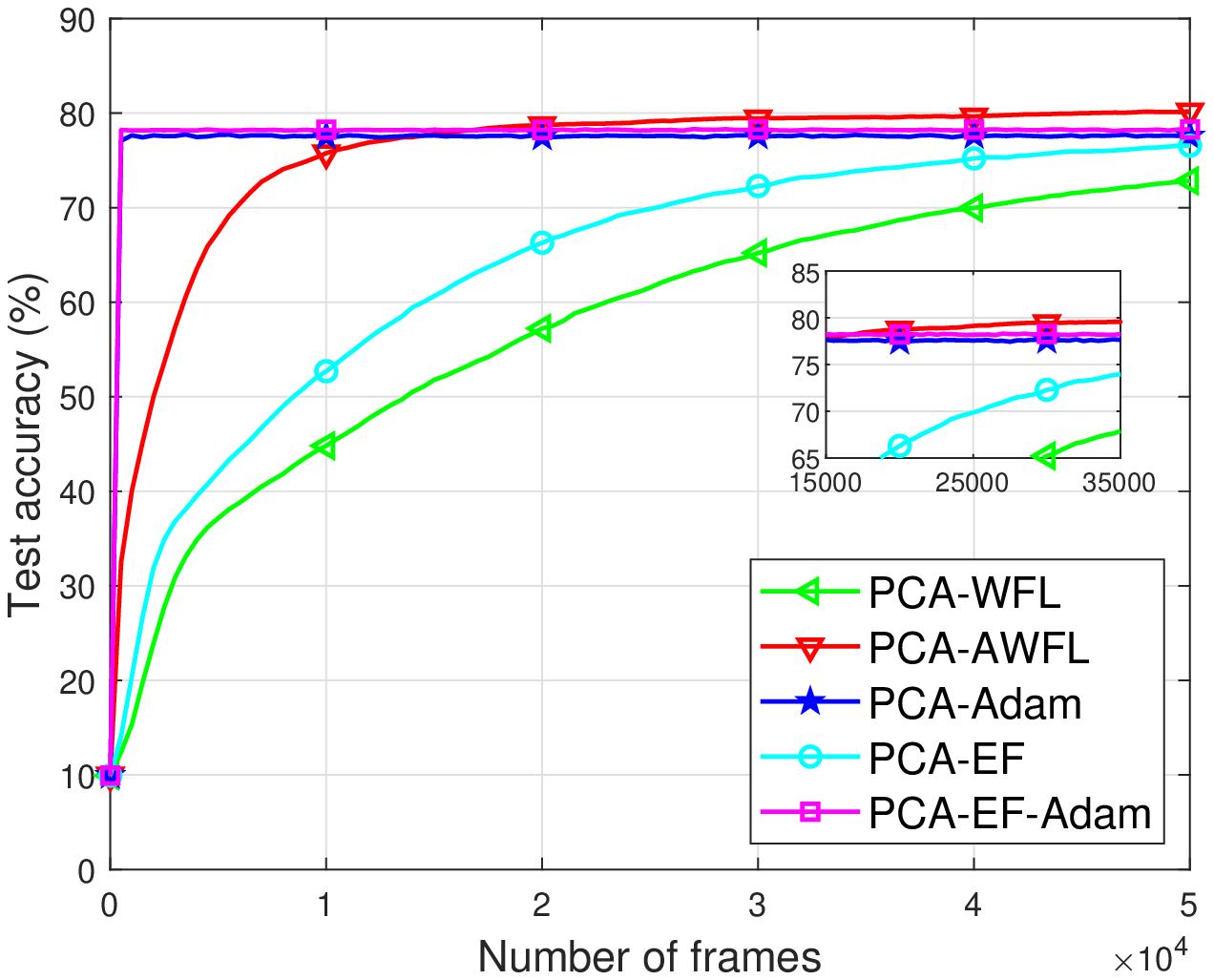}\label{fig:test_acc0001_cifar}}\hspace{0.1cm}
\subfigure[CIFAR-10 with $h_0 = 0.0001$.]{\includegraphics[width=0.3\linewidth]{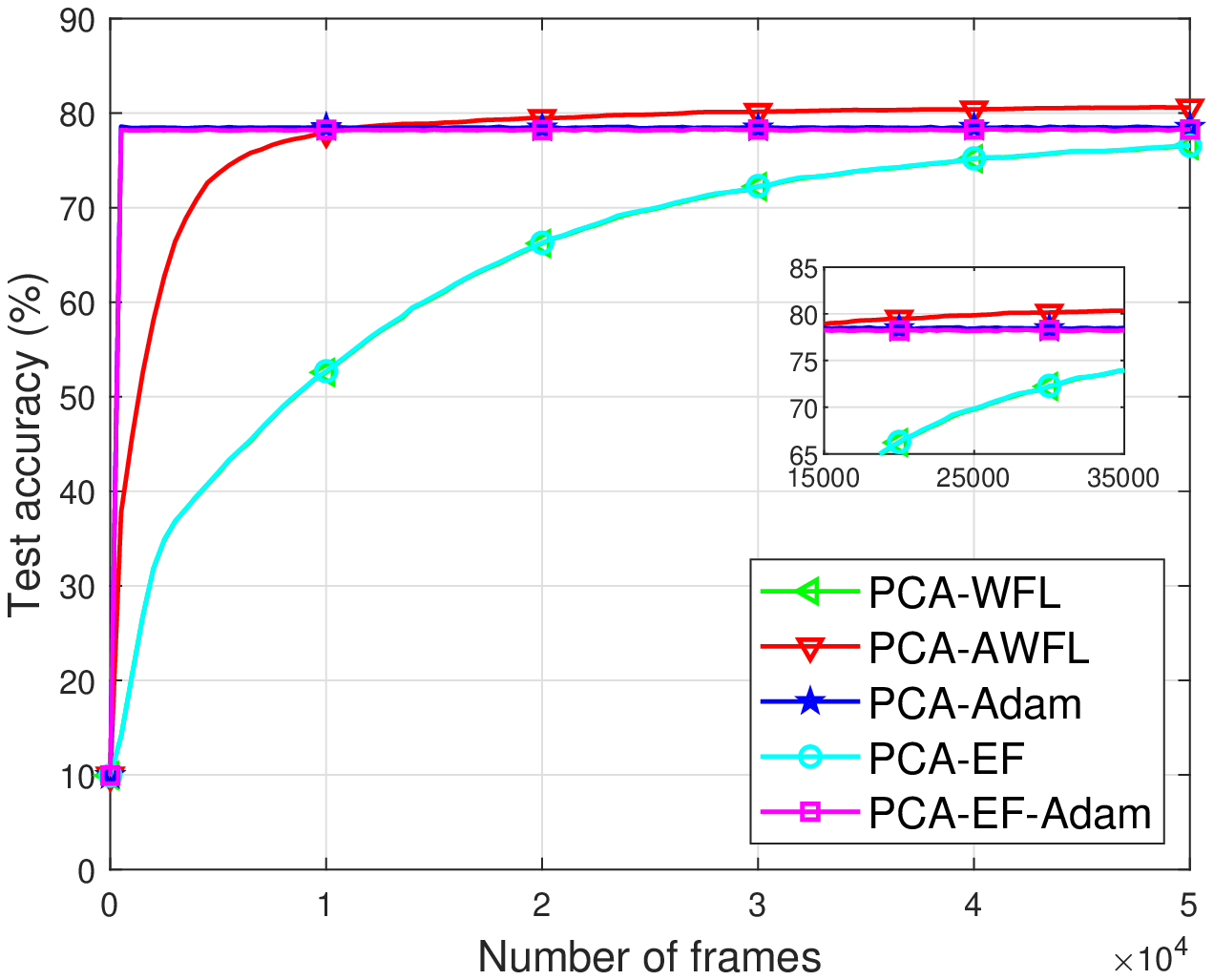}\label{fig:test_acc00001_cifar}}
\subfigure[AWE with $h_0 = 0.001$.]{\includegraphics[width=0.3\linewidth]{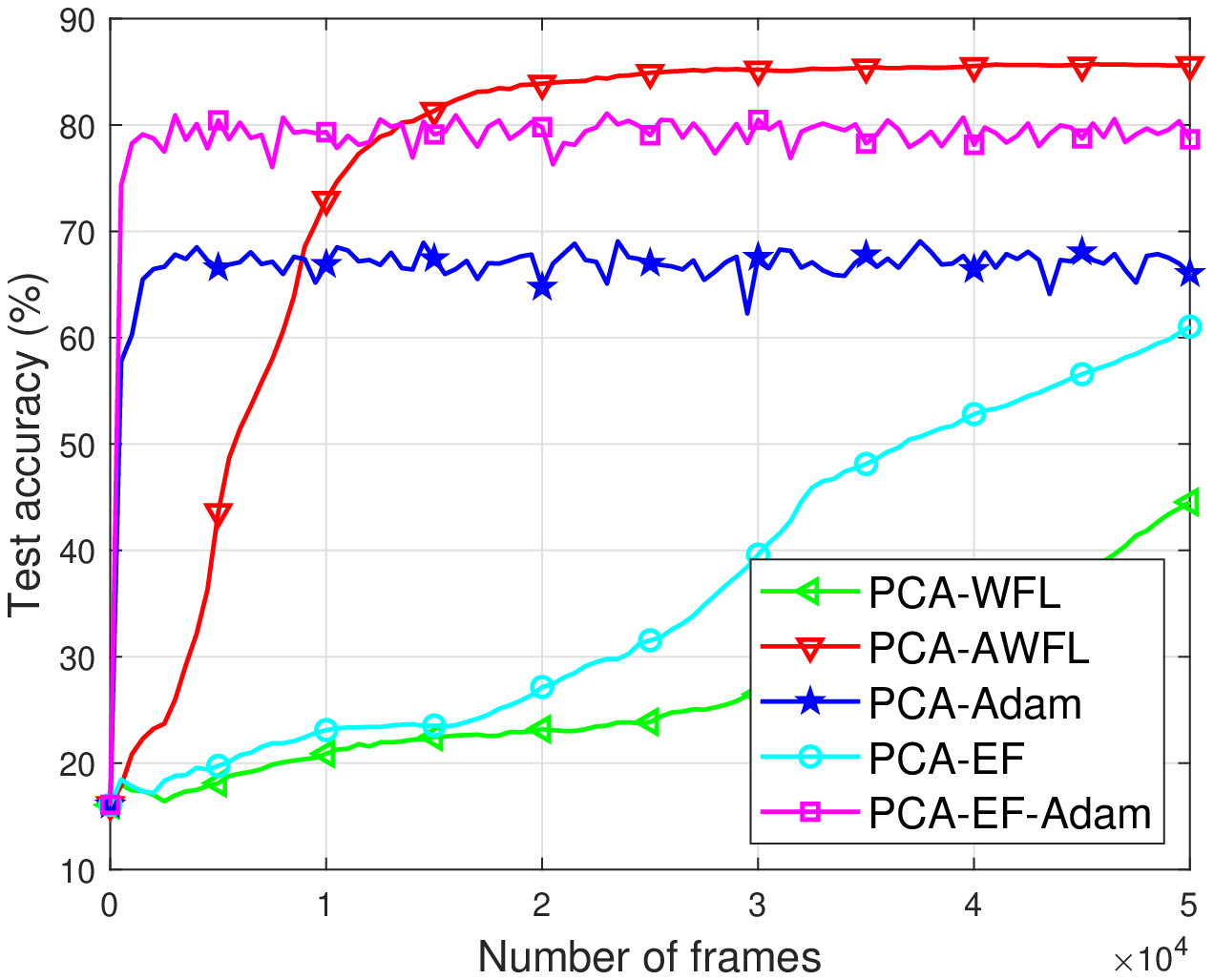}\label{fig:test_acc0001_awe}}\hspace{0.1cm}
\subfigure[AWE with $h_0 = 0.0001$.]{\includegraphics[width=0.3\linewidth]{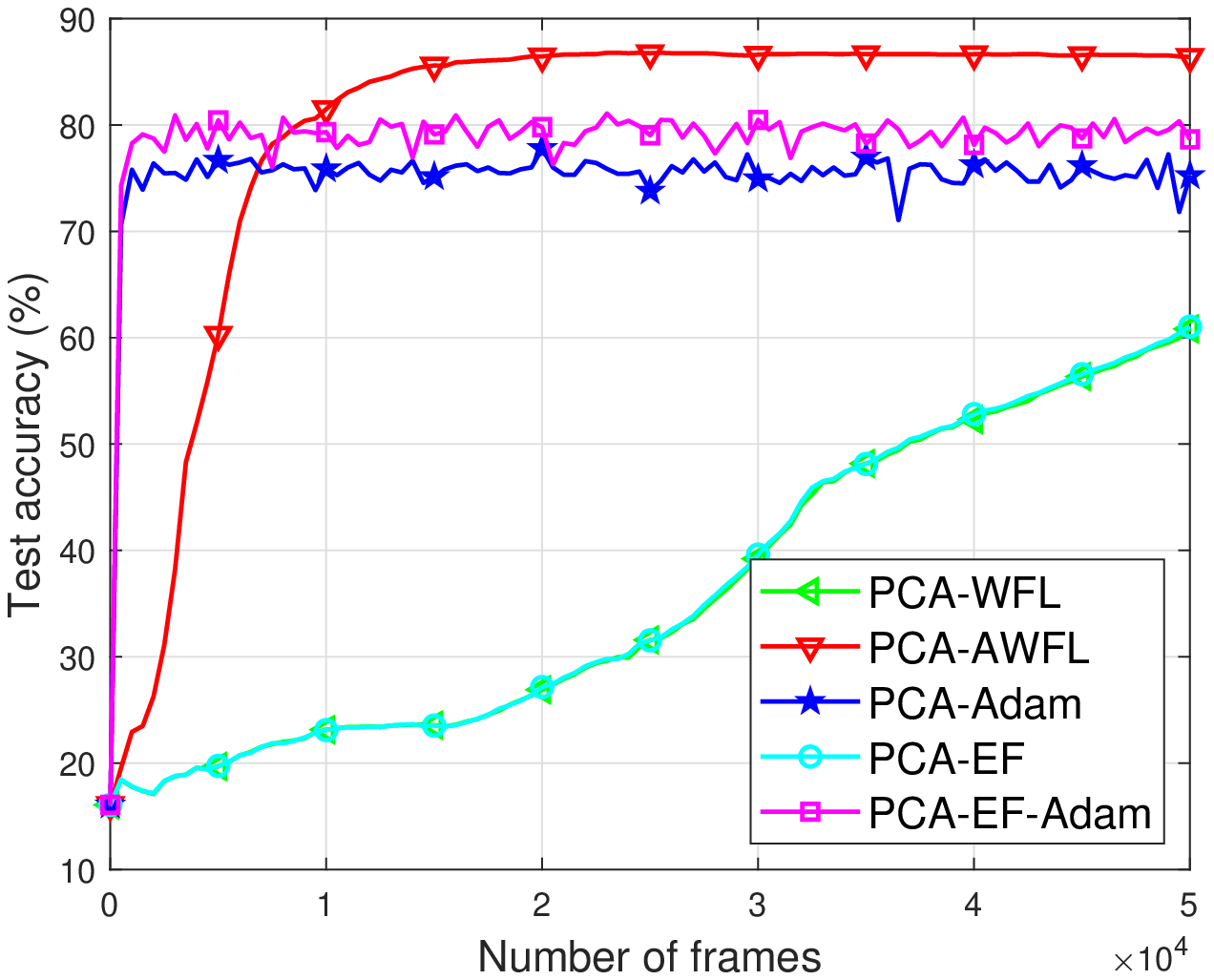}\label{fig:test_acc00001_awe}}
\caption{The convergence behaviours of the testing accuracy for the FMNIST, \mbox{CIFAR-10} and AWE datasets. }\label{fig:initconverge0}
\end{figure}

Figure \ref{fig:initconverge0} shows the convergence behaviours after $5 \times 10^4$ frames for the PCA-WFL, PCA-AWFL, PCA-Adam, PCA-EF, PCA-EF-Adam algorithms for FMNIST, CIFAR-10, and AWE datasets.
From Fig. \ref{fig:initconverge0}, we observe that the PCA-Adam and PCA-EF-Adam achieve the faster convergence rates and the lower testing accuracies than our proposed PCA-AWFL algorithm for the three datasets with $h_0 = 0.001$ and $h_0 = 0.0001$. 
Moreover, from Figs. \ref{fig:test_acc0001_awe} and \ref{fig:test_acc00001_awe}, we observe that the several fluctuations in the testing accuracies of PCA-Adam and PCA-EF-Adam algorithms. 
This is due to the fact that the myopic usage of adaptive gradients induces Adam-type algorithms to stick to some local optimal models with the poorer quality compared with our proposed PCA-AWFL algorithm since the non-convex empirical risk has multiple local optimal models. 
When comparing Figs. \ref{fig:test_acc0001}, \ref{fig:test_acc0001_cifar}, and \ref{fig:test_acc0001_awe} with Figs.  \ref{fig:test_acc00001}, \ref{fig:test_acc00001_cifar}, and \ref{fig:test_acc00001_awe}, we observe that the required frames of our proposed PCA-AWFL algorithm decrease with the truncation threshold $h_0$. 
This is due to the fact that a smaller value of $h_0$ grants the workers to upload more accurate global gradient estimators and thereby reduces the required frames for convergence.  
We also observe that the PCA-AWFL algorithm achieves the highest testing accuracy among the five algorithms. 
This is due to the fact that, in each frame, the PCA-AWFL algorithm can use the historical aggregated gradients that contain more intelligence.
By using more intelligence to adjust the searching direction of the model, the PCA-AWFL algorithm can achieve the best testing accuracy. 

Figure \ref{fig:initconverge0} also shows that the gaps between our proposed PCA-AWFL algorithm and the benchmarks (i.e., \mbox{PCA-EF} and PCA-WFL algorithms) decrease as the number of frames becomes sufficiently large. 
Since the non-convex loss function has multiple local optimal models, we conclude that the model parameters of PCA-EF and PCA-WFL algorithms can converge to the neighbor of the local optimal models that have similar testing accuracy as the PCA-AWFL algorithm. 

The convergence of gradient norms (in Fig. \ref{fig:grad_norm}) and testing accuracies (in Fig. \ref{fig:initconverge0}) jointly verify our envision: the testing accuracies of \mbox{PCA-WFL} and \mbox{PCA-AWFL} algorithms reach the peaks when the intelligence stops spreading.

\begin{figure}[ht]
\centering
\subfigure[FMNIST dataset.]{\includegraphics[width= 0.33\linewidth]{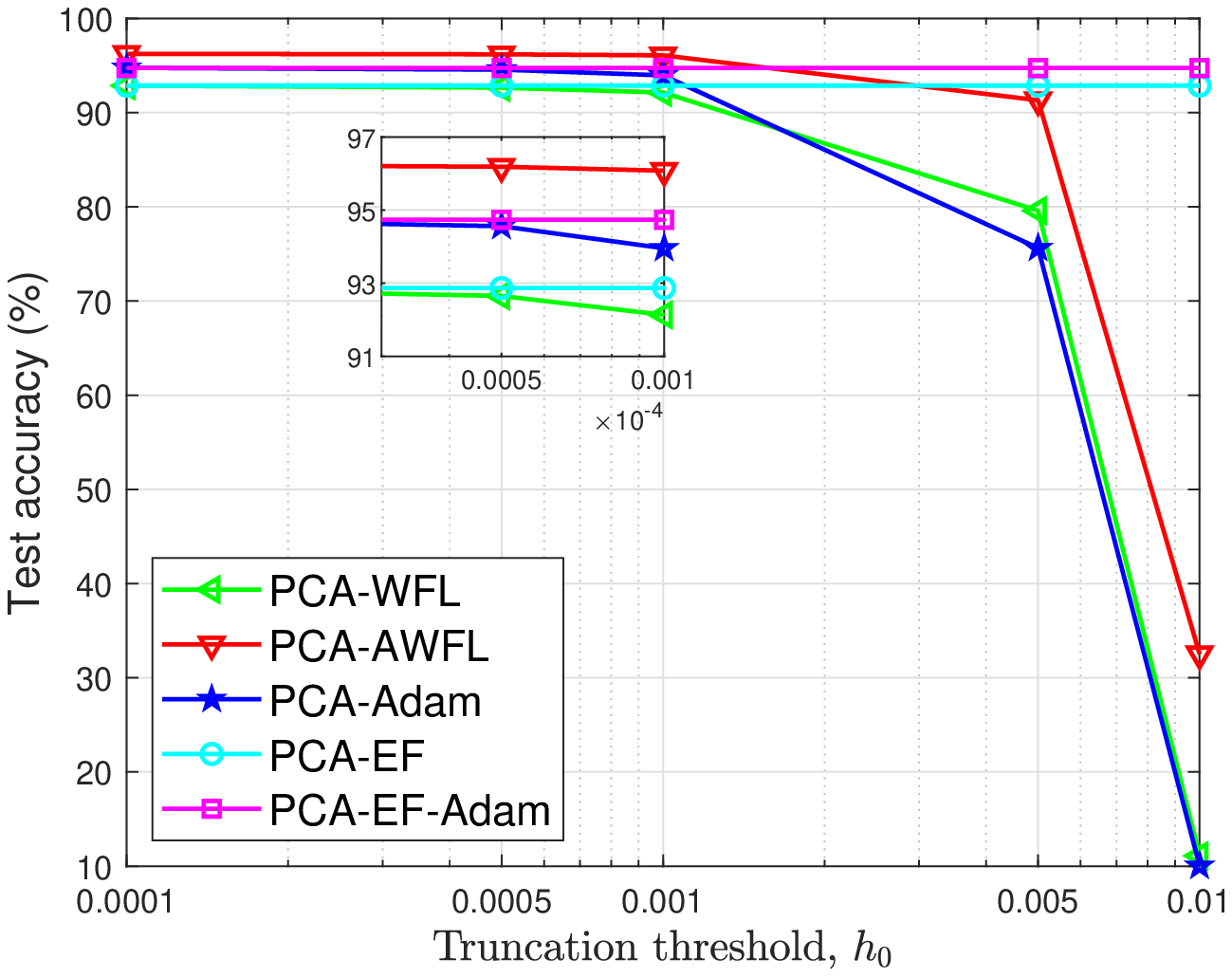}}\hspace{-0.3 cm}
\subfigure[CIFAR-10 dataset.]{\includegraphics[width= 0.33\linewidth]{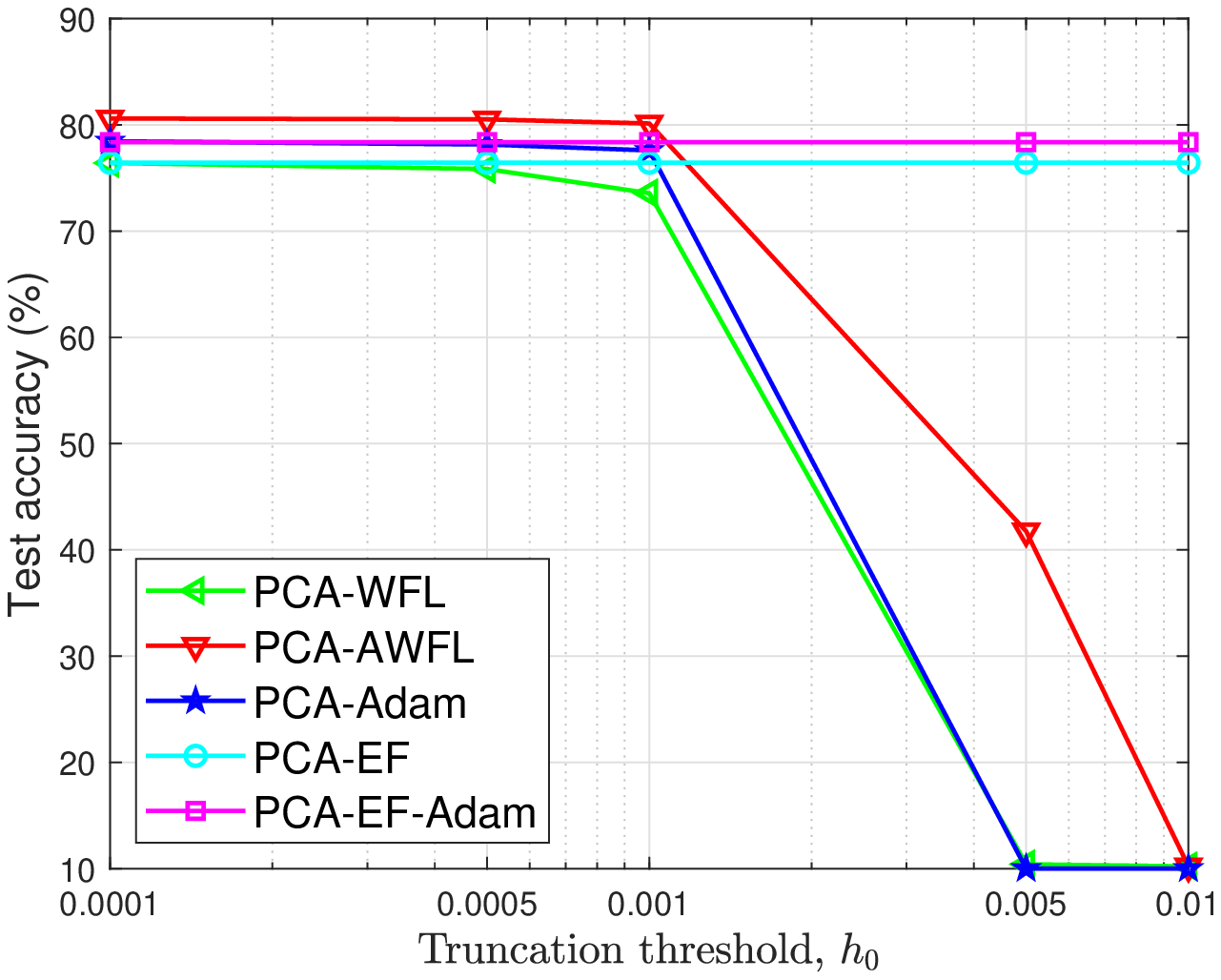}}\hspace{-0.3 cm}
\subfigure[AWE dataset.]{\includegraphics[width= 0.33\linewidth]{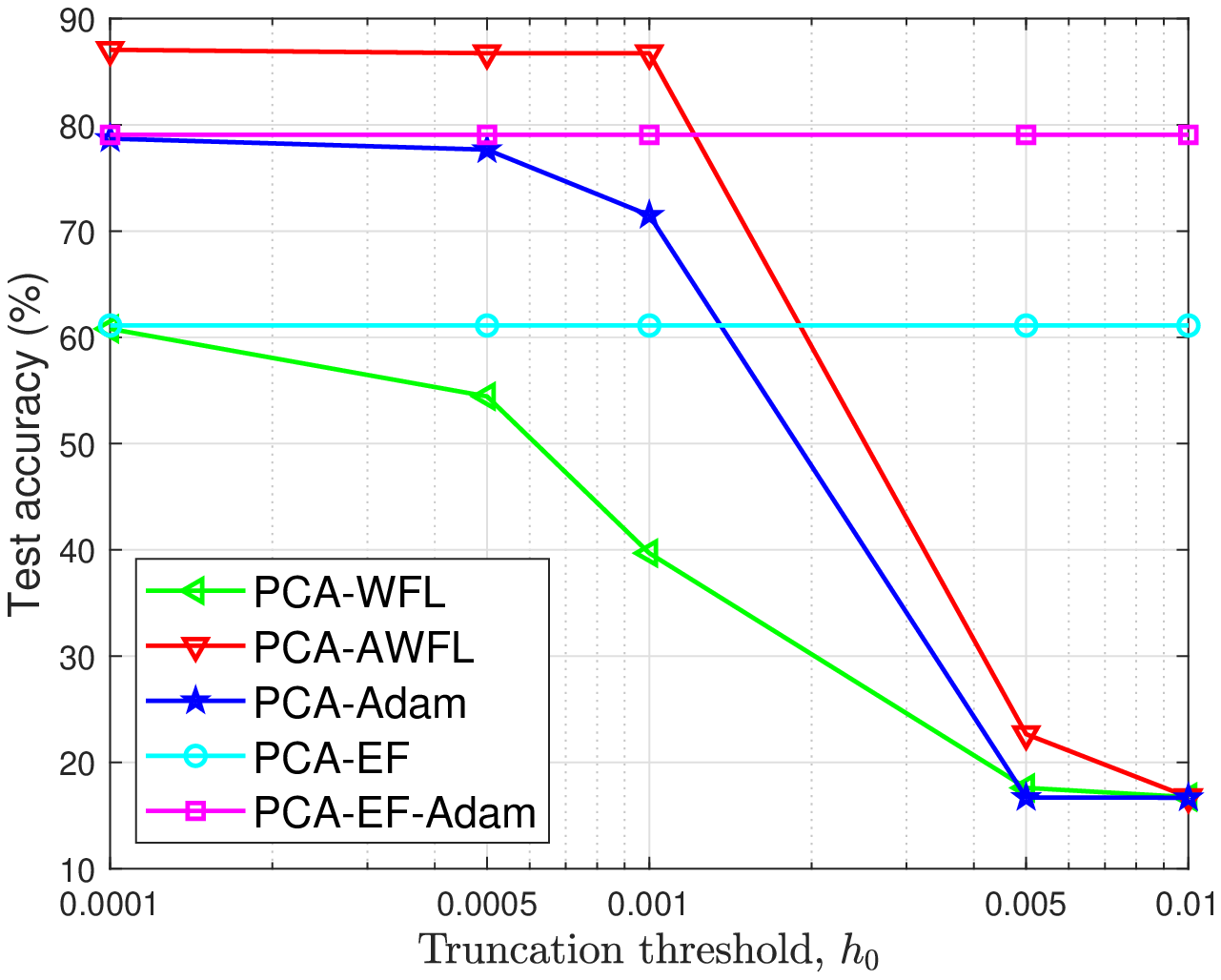}}
\caption{Testing accuracy under different truncation thresholds for the FMNIST, CIFAR-10, and AWE datasets.}
\label{fig:initthreshold}
\subfigure[FMNIST dataset.]{\includegraphics[width= 0.33\linewidth]{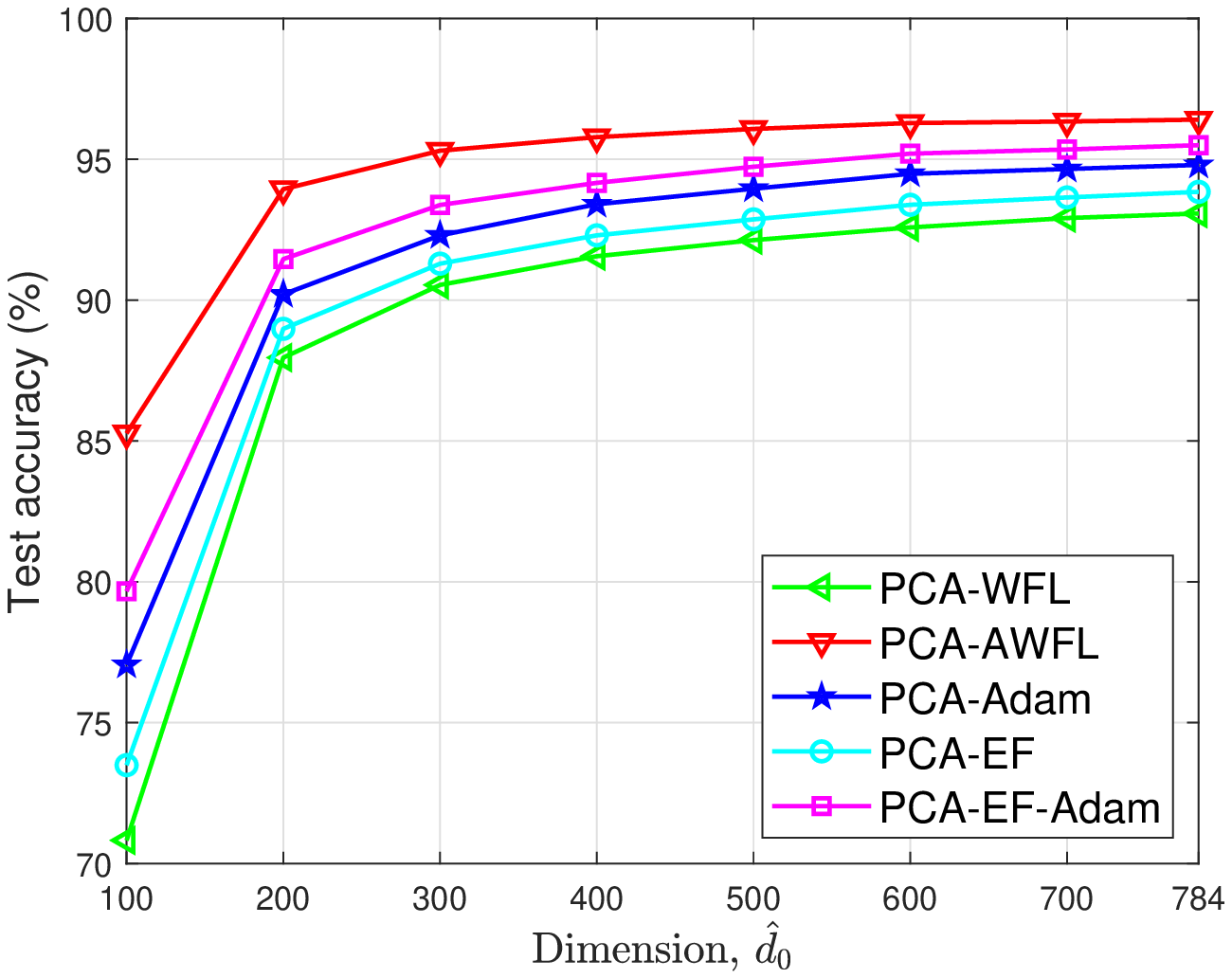}\label{fig:dim_acc_avg_fmnist}}\hspace{-0.3 cm}
\subfigure[CIFAR-10 dataset.]{\includegraphics[width= 0.33\linewidth]{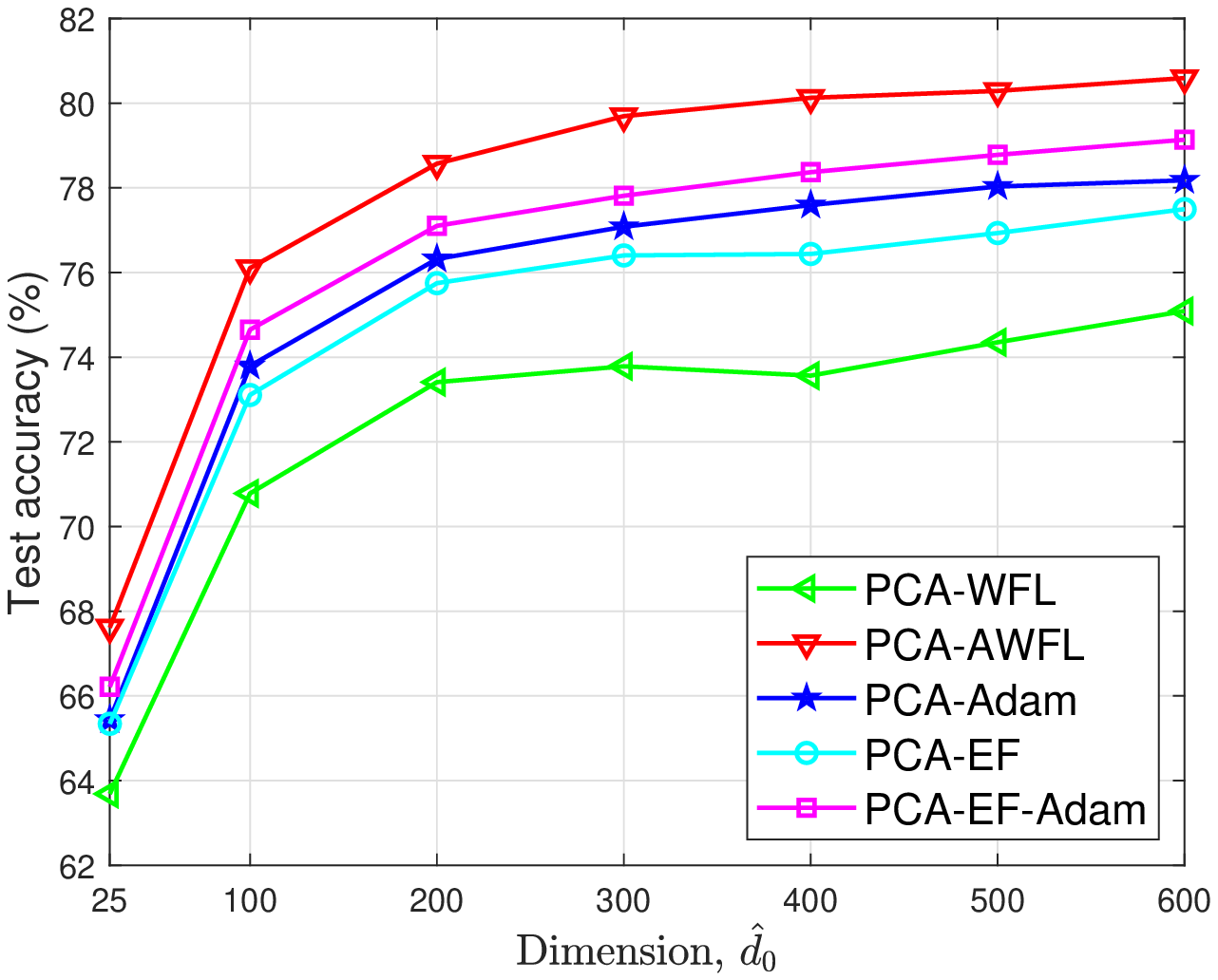}\label{fig:dim_acc_avg_cifar}}\hspace{-0.3 cm}
\subfigure[AWE dataset.]{\includegraphics[width= 0.33\linewidth]{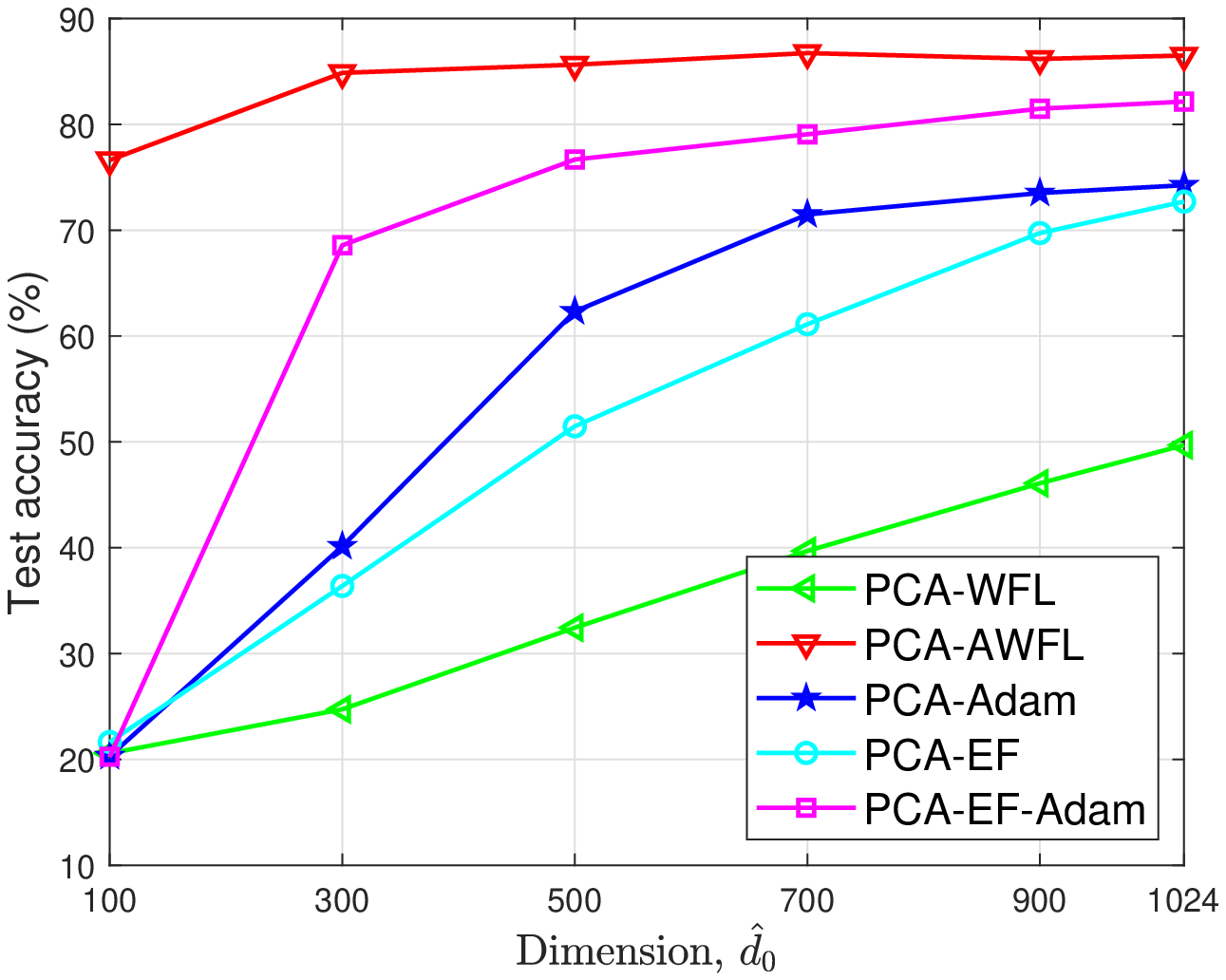}\label{fig:dim_acc_avg_awe}}
\caption{Testing accuracy v.s. the dimension of data sample $\hat d_0$  for the FMNIST, CIFAR-10, and AWE datasets.}
\label{fig:dim_acc_avg}
\end{figure}

Figure \ref{fig:initthreshold} shows the impacts of truncation threshold $h_0$ on the testing accuracies for the PCA-WFL, PCA-AWFL, PCA-Adam, PCA-EF, PCA-EF-Adam algorithms for FMNIST, CIFAR-10, and AWE datasets.
We observe that our proposed PCA-AWFL algorithm achieves the highest testing accuracy under different truncation thresholds. 
Note that the PCA-EF and PCA-EF-Adam algorithms are insensitive to the truncation threshold since the local gradients are assumed to be uploaded over error-free channels. 
We observe that the accuracies of the PCA-WFL, PCA-AWFL, PCA-Adam algorithms monotonically decrease with the truncation threshold $h_0$. 
This is due to the fact that a larger truncation threshold $h_0$ indicates a smaller chance of uploading local gradients such that a less accurate estimator of the global gradient is obtained at the server.
A less accurate estimator of global gradient leads to a lower testing accuracy under the given number of frames.

Figure \ref{fig:dim_acc_avg} shows the impacts of the one-shot distributed PCA on the testing accuracies of the PCA-WFL, PCA-AWFL, PCA-Adam, PCA-EF, PCA-EF-Adam algorithms for the FMNIST, CIFAR-10, and AWE datasets.
Note that we select the reduced dimension $\hat d_0$ based on the numerical tests in Fig. \ref{fig:dim_acc_avg}.
More specifically, we observe from Fig. \ref{fig:dim_acc_avg_fmnist} that reducing the sample dimension from $784$ to $500$ only sacrifices $0.34\%$ testing accuracy for our proposed PCA-AWFL algorithm over the FMNIST dataset. 
The similar conclusions can be respectively obtained for the CIFAR-10 and AWE datasets from \ref{fig:dim_acc_avg_cifar} and \ref{fig:dim_acc_avg_awe}. 
Recalling Section \ref{sec:pca}, we conclude that our proposed PCA-AWFL algorithm achieves a near optimal testing accuracy with communication expenditure reduced by $500/784 \approx 36.22\%$. }

%We observe that the testing accuracy of the error-free gradient descent, PCA-WFL, and PCA-AWFL algorithms monotonically increases  with the dimension of reduced sample $\hat d_0$, while the accuracy of the sparse gradient descent algorithm is insensitive to the dimension of reduced sample. 
%This is due to the facts: 1) the error-free gradient descent, PCA-WFL, and PCA-AWFL algorithms can use more features to gain a more accurate training model; and 2) the sparse gradient descent algorithm leverages an accumulated error compensate method to correct the sparsity of local gradients. 

\section{Concluding Remarks}
This work investigated the analog WFL system, where the workers have non-i.i.d. locations and upload local gradients to the server via orthogonal wireless channels.
A one-shot distributed PCA had been used to reduce the dimension of local gradients, and the channel scheduling probability had been included in the truncated channel inversion to obtain an unbiased estimator of the aggregated gradient. 
Based on the one-shot PCA, truncated channel inversion and the Nesterov's momentum, the PCA-WFL and \mbox{PCA-AWFL} algorithms had been proposed.
A finite-time analysis had been performed to quantify the convergence behaviours of the PCA-WFL and  \mbox{PCA-AWFL} algorithms. 
Leveraging the intelligence networking, an intuition had been given to illustrate the existence of the speedup capability of Nesterov's momentum. 
Besides, the finite-time analysis had shown that the PCA-WFL and  \mbox{PCA-AWFL}  algorithms can achieve the parallel speedup with respect to the number of workers. 
{\color{blue}
Our proposed PCA-AWFL algorithm: 1) is more robust to the channel fadings and noise compared with the PCA-EF and \mbox{PCA-WFL} algorithms; 2) attains a higher testing accuracy compared with the state-of-the-art \mbox{PCA-EF-Adam} and \mbox{PCA-Adam} algorithms; 3) can be adapted to more practical datasets; 4) has a lower computational complexity per frame compared with the \mbox{PCA-EF-Adam} and \mbox{PCA-Adam} algorithms.
The only disadvantage of our proposed PCA-AWFL algorithm is the slower convergence behaviour compared with the PCA-Adam and PCA-EF-Adam algorithms over three datasets.}

\appendices
\section{Proof of Lemma \ref{le:02}}\label{apdx:02}
We first introduce two auxiliary inequalities before proceeding to prove the main results. 
The expectations of $c_{n,k}\one_{n,k} \odot y_{n,k}$ and $\|c_{n,k}\one_{n,k} \odot y_{n,k} - y_{n,k}\|^2$ are, respectively, derived as 
\begin{equation}\label{apdx02:01}
\E{ c_{n,k}\one_{n,k} \odot y_{n,k} } = c_{n,k} \E{ \one_{n,k} } \odot y_{n,k} = y_{n,k}
\end{equation}
and 
\begin{equation}\label{apdx02:02}
\E{ \|c_{n,k}\one_{n,k} \odot y_{n,k} - y_{n,k}\|^2 } = (c_{n,k} - 1)\|y_{n,k}\|^2. 
\end{equation}

We also introduce a useful expectation $\E{ \rho_{n,k}^{-2} }$ (See Appendix \ref{apdx:01} for the detailed derivations) as 
\begin{equation}\label{apdx02:02a}
\E{ \rho_{n,k}^{-2} } = p_0^{-1} \delta_{n,k}^{\alpha} \exp(-2\delta_{n,k}^{\alpha} h_0^2) E_1( \delta_{n,k}^{\alpha}h_0^2 ).
\end{equation}

Recalling the condition $\frac{1}{N}\sum_{n=1}^N\|y_{n,k} - \nabla f(w_k)\|^2 \le G$ and using the fact $\|x + y\|^2 \le 2\|x\|^2 + 2 \|y\|^2$, we have 
\begin{equation}\label{apdx02:03}
\frac{1}{N}\sum_{n=1}^N \|y_{n,k}\|^2 \le 2G + 2\|\nabla f(w_k)\|^2.
\end{equation}

Based on \eqref{apdx02:01} and the zero-mean channel noise, we obtain $\E{ \frac{1}{N}\sum_{n=1}^N \hat \nabla_{n,k} } = \frac{1}{N}\sum_{n=1}^N y_{n,k}$.

We derive the upper bound of the expectation of $\|\frac{1}{N}\sum_{n=1}^N \one_{n,k} \odot y_{n,k} \|^2$ as 
\begin{subequations}\label{apdx02:04}
\begin{align}
&\EE{ \| \frac{1}{N} \sum_{n=1}^N c_{n,k} \one_{n,k} \odot y_{n,k}\|^2  } \label{apdx02:04a}\\
&\!\!=\! \EE{ \| \frac{1}{N} \sum_{n=1}^N\! ( c_{n,k} \one_{n,k} \odot y_{n,k} - y_{n,k} + y_{n,k}) \|^2 } \label{apdx02:04b}\\
&\!\!=\! \EE{ \|\frac{1}{N} \sum_{n=1}^N\! ( c_{n,k} \one_{n,k} \odot y_{n,k} \!-\! y_{n,k})\|^2 } \!+\!  \|\nabla f(w_k)\|^2  \label{apdx02:04c} \\
&\!\!=\! \frac{1}{N^2}\sum_{n=1}^N\!  \E{ \| c_{n,k} \one_{n,k} \odot y_{n,k} \!-\! y_{n,k} \|^2 } \!+\!  \|\nabla f(w_k)\|^2 \label{apdx02:04d} \\
&\!\!=\! \frac{1}{N^2}\sum_{n=1}^N (c_{n,k} - 1)\|y_{n,k}\|^2 + \|\nabla f(w_k)\|^2 \label{apdx02:04e}\\
&\!\!\le\!  \frac{c_1}{2N} +\frac{1}{2N^2}\sum_{n=1}^N \|y_{n,k}\|^2 + \|\nabla f(w_k)\|^2 \label{apdx02:04f}
%&\!\!\le\!  \sq{1 + \frac{a_2}{2N(1-a_1)}} \|\nabla f(w_k)\|^2 + \frac{c_1}{2a_2N} + \frac{a_2 G}{2 a_1 N} \label{apdx02:04g}
\end{align}
\end{subequations} 
where the equality \eqref{apdx02:04c} is based on  $\E{ \|X\|^2 } = \E{ \|X - \E{X}\|^2 } + \|\E{ X} \|^2$ and $\nabla f(w_k) = \frac{1}{N} \sum_{n=1}^N  y_{n,k}$;
the equality  \eqref{apdx02:04d} follows from the fact that the vectors $c_{n,k}\one_{n,k} \odot y_{n,k} - y_{n,k}$ are independent for workers with $\E{  c_{n,k}\one_{n,k} \odot y_{n,k} \!-\! y_{n,k} } = 0$; 
the equality \eqref{apdx02:04e} is based on \eqref{apdx02:02};
and, the inequality \eqref{apdx02:04f} follows from $\inp{x, y} \le \frac{1}{2}\|x\|^2 + \frac{1}{2}\|y\|^2$ and the triangle inequality with $c_1 := \frac{1}{N} \sum_{n=1}^N (c_{n,k} \!-\! 1)$.

The derivation for the upper bound of accumulated noise is as follows. 
We take expectation over the squared norm of accumulated noise per frame $k$ as 
\begin{subequations}\label{apdx02:05}
\begin{align}
\E{ \| z_k \|^2 } 
&= \frac{1}{N^2} \sum_{n=1}^N \EE{ 
	\frac{ \| y_{n,k} \|^2 }{\rho_{n,k}^2} \|z_{n,k}\|^2 } \label{apdx02:05c} \\
&\le \frac{1}{N^2}  \sum_{n=1}^N d_1 \sigma^2 \EE{ \frac{ \| y_{n,k} \|^2 }{\rho_{n,k}^2} }  \label{apdx02:05d} \\
&\le \frac{1}{2N^2}\sum_{n=1}^N \|y_{n,k}\|^2 + \frac{c_2 d_1\sigma^2 }{2N p_0} \label{apdx02:05e} 
% \label{apdx02:05f}
\end{align} 
\end{subequations}
where equality \eqref{apdx02:05c} follows from the fact that the additive white Gaussian noise vectors  $[z_{n,k}]_{n=1}^N$ are  i.i.d. with mean zero and covariance $\sigma^2 I$;
the inequality \eqref{apdx02:05d} follows from the fact that channel noise is independent of channel fadings and the location of workers with $\E{  \|z_{n,k}\|^2 } = d_1 \sigma^2$; 
the inequality \eqref{apdx02:05e} is obtained by $\inp{x, y} \le \frac{1}{2}\|x\|^2 + \frac{1}{2}\|y\|^2$ and the triangle inequality with $c_2 := \frac{1}{N p_0} \sum_{n=1}^N \delta_{n,k}^{\alpha} \exp(2\delta_{n,k}^{\alpha} h_0^2) E_1( \delta_{n,k}^{\alpha}h_0^2 )$.

Note that the channel noise $z_k$ is independent of the channel fadings per frame $k$ and the channel fadings, we have 
\begin{subequations}\label{apdx02:06}
\begin{align}
&\EE{ \| \frac{1}{N}\sum_{n=1}^N \hat \nabla_{n,k} \|^2 } \label{apdx02:06a}\\
&\le \EE{ \|\frac{1}{N}\sum_{n=1}^N c_{n,k}\one_{n,k} \odot y_{n,k} \|^2 } + \E{ \| z_k \|^2 }   \label{apdx02:06b} \\
&\le \frac{1}{N^2}\sum_{n=1}^N \|y_{n,k}\|^2 +  \frac{c_1}{2N} +  \frac{c_2 d_1\sigma^2 }{2N} + \|\nabla f(w_k)\|^2 \label{apdx02:06c} \\
&\le 2\|\nabla f(w_k)\|^2  + \frac{c_1 + c_2 d_1 p_0^{-1}\sigma^2 + 4G}{ 2N} \label{apdx02:06d}
\end{align}
\end{subequations}
where the inequality \eqref{apdx02:06c} the facts in \eqref{apdx02:04f} and \eqref{apdx02:05e}; and \eqref{apdx02:06d} follows from \eqref{apdx02:03}  with $N \ge 2$.

\section{Derivation of \eqref{apdx02:02a}}\label{apdx:01}
When the power of channel coefficient satisfies $|h_{n,k}[i]| \ge h_0$, the channel $i$ is scheduled for worker $n$ per frame $k$. 
Following the channel alignment policy in \eqref{eqa:04}, we have
\begin{equation}\label{apdx01:01}
\frac{ |\one_{n,k}[i]y_{n,k}[i]|^2 }{|h_{n,k}[i]|^2\|y_{n,k}\|^2} \!=\! \left\{
	\begin{matrix}
		\! \frac{|y_{n,k}[i]|^2}{|h_{n,k}[i]|^2 \| y_{n,k} \|^2}, &\!\!  |h_{n,k}[i]| \ge h_0\\
		\! 0, &\!\! |h_{n,k}[i]| < h_0. 
	\end{matrix}
	\right.
\end{equation}

Recalling the fact that each $h_{n,k}[i]$ follows i.i.d. circularly symmetric complex Gaussian distribution ${\cal CN}(0, \delta_{n,k}^{-\alpha})$. 
The power of each $|h_{n,k}[i]|^2$ is i.i.d. exponentially distributed with probability density function as $\delta_{n,k}^{\alpha}\exp(-\delta_{n,k}^{\alpha} x)$, $\forall n$. 

Expanding the term $\| p_{n,k}  \odot \frac{y_{n,k}}{\|y_{n,k}\|} \|^2$ and leveraging the power budget constraint \eqref{eqa:03}, we obtain
\begin{equation}\label{apdx01:02aa}
\rho_{n,k}^{-2} = \frac{c_{n,k}^2}{p_0 } \sum_{i=1}^{d_1} \frac{ |\one_{n,k}[i]y_{n,k}[i]|^2 }{|h_{n,k}[i]|^2\|y_{n,k}\|^2}.
\end{equation}

Given the gradient $y_{n,k}$ and location $\delta_{n,k}$, we have
\begin{subequations}\label{apdx01:02}
	\begin{align}
		 \E{ \rho_{n,k}^{-2}  }  
		&= \frac{c_{n,k}^2}{p_0 } \EE{  \sum_{i=1}^d \frac{ |\one_{n,k}[i]y_{n,k}[i]|^2 }{|h_{n,k}[i]|^2\|y_{n,k}\|^2}  \Big|  y_{n,k}, \delta_{n,k} } \label{apdx01:02b}\\
		&=\frac{c_{n,k}^2}{p_0 } \sum_{i=1}^{d_1}  \frac{ |y_{n,k}[i]|^2 }{ \| y_{n,k} \|^2} \int_{h_0^2}^{\infty} \frac{ \delta_{n,k}^{\alpha} }{x \exp(\delta_{n,k}^{\alpha} x)}    d x  \label{apdx01:02c}\\
		&= \frac{c_{n,k}^2}{p_0 } \sum_{i=1}^{d_1}  \frac{|y_{n,k}[i]|^2}{ \| y_{n,k} \|^2} \delta_{n,k}^{\alpha} E_1( \delta_{n,k}^{\alpha}h_0^2 ) \label{apdx01:02d}\\
		&= p_0^{-1} \delta_{n,k}^{\alpha} \exp(2\delta_{n,k}^{\alpha} h_0^2) E_1( \delta_{n,k}^{\alpha}h_0^2 ) \label{apdx01:02e}
	\end{align}
\end{subequations}
where the inequality in \eqref{apdx01:02c} follows from \eqref{apdx01:01};  
the equality in \eqref{apdx01:02d} follows from the definition of the exponential integral of $E_1(\cdot)$ in  \cite[pp. xxxv]{Jeffrey2007};
and, the equality in \eqref{apdx01:02e} follows from the facts $\sum_{i=1}^{d_1} \nicefrac{ |y_{n,k}[i]|^2 }{ \| y_{n,k} \|^2} = 1$ and $c_{n,k} = \exp(\delta_{n,k}^{\alpha} h_0^2)$.

\section{Proof of Theorem \ref{th:vanilla}}\label{apdx:03}
We start the proof by using the $L$-Lipschitz continuous condition as 
\begin{equation}\label{apdx03:01}
\!\!\!\! f(w_{k+1}) \!-\! f(w_k) 
\!\!\le\!\! \inp{ \nabla\! f(w_k), w_{k+1} \!\!-\! \! w_k } \!+\!\! \frac{L}{2}\| w_{k+1} \!\!-\!\!  w_{k} \|^2\!. \!\!\!\!
\end{equation}

Based on the recursion \eqref{eqa:11}, we have $\inp{ \nabla f(w_k), w_{k+1} - w_k } = -\eta \inp{\nabla f(w_k), \frac{1}{N}\! \sum_{n=1}^N\!\hat \nabla_{n,k}}$. 
Taking expectation of $\inp{\nabla f(w_k), \frac{1}{N}\! \sum_{n=1}^N\!\hat \nabla_{n,k}}$ over the randomness of wireless channels and the locations of workers, we have
\begin{subequations}\label{apdx03:03}
\begin{align}
& \E{ \inp{ \nabla f(w_k), \frac{1}{N}\! \sum_{n=1}^N\!\hat \nabla_{n,k} } } = \E{ \| \nabla f(w_k)\|^2 }. \label{apdx03:03c}
\end{align}
\end{subequations}
where the equality \eqref{apdx03:03c} follows from the facts $\E{\E{X}} = \E{X}$ and Lemma \ref{le:02}.

Therefore, the expectation of the first term on the RHS of \eqref{apdx03:01} is
\begin{equation}\label{apdx03:04}
\E{ \inp{ \nabla f(w_k), w_{k+1} - w_k } } = -\eta\E{ \| \nabla f(w_k)\|^2 }. 
\end{equation}

The expectation of the second term on the RHS of \eqref{apdx03:01} is derived as 
\begin{subequations}\label{apdx03:05}
\begin{align}
&\EE{ \frac{L}{2}\| w_{k+1} -  w_{k} \|^2 } \\
&= \frac{\eta^2L}{2}\EE{ \|\frac{1}{N}\! \sum_{n=1}^N\!\hat \nabla_{n,k}\|^2 } \label{apdx03:05a}\\
&\le \eta^2 L \|\nabla f(w_k)\|^2 + \frac{c_1 + c_2 d_1 p_0^{-1}\sigma^2 + G}{4N}\eta^2 L \label{apdx03:05b} 
\end{align}
\end{subequations}
where the equality \eqref{apdx03:05b} follows from the recursion \eqref{eqa:11}; and the inequality \eqref{apdx03:05b} follows from Lemma \ref{le:02}. 

Substituting \eqref{apdx03:04} and \eqref{apdx03:05b} into the expectation of \eqref{apdx03:01}, we obtain
\begin{equation}\label{apdx03:06}
\begin{split}
& \E{ f(w_{k+1}) - f(w_k)  } \\
&\le \eta (\eta L \!-\! 1)\E{ \|\nabla f(w_k)\|^2 } \!+\! \frac{c_1 \!+\! c_2 d_1 p_0^{-1} \sigma^2 \!+\! G}{4N} \eta^2 L. 
\end{split}
\end{equation}

Dividing both sides of \eqref{apdx03:06} by $\eta K$ and summing over $k = 0, \ldots, K-1$, we have
\begin{equation}\label{apdx03:07}
\begin{split}
&\frac{1}{\eta K}\E{ f(w_{K}) - f(w_0)  } \\
&\le \frac{\eta L \!-\! 1}{K}\!\sum_{k=0}^{K-1}\! \E{ \|\nabla f(w_k)\|^2 }   \!+\! \frac{c_1 \!+\! c_2 d_1 p_0^{-1} \sigma^2 \!+\! G}{4N}\eta L. 
\end{split}
\end{equation}

Setting $\eta \le  \nicefrac{3}{4 L}$, we have $1 - \eta L \ge \frac{1}{4}$. 
Using the fact $f(x_K) \ge f^*$, we obtain
\begin{equation}\label{apdx03:08}
\begin{split}
&\frac{1}{K} \!\! \sum_{k=0}^{K-1}\!\! \E{ \| \nabla f(w_k) \|^2 }  \\
&\le\! \frac{ 4 }{\eta K} \E{ f(w_{0}) -\! f^* }  +  \frac{c_1 \!+\! c_2 d_1 p_0^{-1} \sigma^2 \!+\! G}{N}\eta L. 
\end{split}
\end{equation}

\section{Proof of Theorem \ref{th:01}}\label{apdx:04}
Before proceeding the proof, we set $x_0 = w_0$ and define an auxiliary sequence with $k \ge 1$ as
\begin{equation}\label{apdx04:01}
x_k = \frac{1}{ 1-\beta }w_k - \frac{\beta}{1-\beta} w_{k-1} + \frac{\eta\beta}{1-\beta} \frac{1}{N}\sum_{n=1}^N  \hat \nabla_{n,k-1}.
\end{equation}

Following several algebraic manipulations, we have
\begin{equation}\label{apdx04:02}
x_{k+1} - x_k = - \frac{\eta}{1 - \beta} \frac{1}{N} \sum_{n=1}^N \hat \nabla_{n,k}, k \ge 0
\end{equation}
and 
\begin{equation}\label{apdx04:03}
x_{k} - w_{k} = -\frac{\eta \beta^2}{1-\beta} u_k, k \ge 0.
\end{equation}

Recalling \eqref{apdx04:04} and summing $\|u_k\|^2$ over $k = 0, \ldots, K-1$, we have 
\begin{subequations}\label{apdx04:05}
\begin{align}
\sum_{k=0}^{K-1} \| u_k \|^2 &= \sum_{k=1}^{K-1} \Big\| \sum_{t = 0}^{k - 1} \beta^{k-1-t} \frac{1}{N}\sum_{n=1}^N \hat\nabla_{n,t} \Big\|^2 \label{apdx04:05a}\\
&= \sum_{k=1}^{K-1} s_k^2   \Big\| \sum_{t = 0}^{k - 1} \frac{\beta^{k-1-t}}{s_k} \frac{1}{N}\sum_{n=1}^N \hat\nabla_{n,t} \Big\|^2  \label{apdx04:05b}\\
&\le \sum_{k=1}^{K-1} \sum_{t = 0}^{k - 1} s_k \beta^{k-1-t} \| \frac{1}{N}\sum_{n=1}^N \hat\nabla_{n,t} \|^2 \label{apdx04:05c} \\ 
&\le \frac{1}{(1-\beta)^2}  \sum_{t=0}^{K-1} \| \frac{1}{N}\sum_{n=1}^N \hat\nabla_{n,t} \|^2  \label{apdx04:05d} 
\end{align}
\end{subequations}
where  $s_k := \sum_{t = 0}^{k-1} \beta^{k-1-t} = \frac{1 - \beta^k}{1 - \beta}$ in \eqref{apdx04:05b}; 
inequality \eqref{apdx04:05c} follows from the convexity of $\ell_2$-norm; 
inequality \eqref{apdx04:05d} follows from $\sum_{k = t+1}^{K-1} s_k \beta^{k-1-t} \le \frac{1}{ (1-\beta)^2 }$. 
	
Using the \mbox{$L$-Lipschitz} continuous property in Condition \ref{cd:02}, we have  
\begin{equation}\label{apdx04:06}
\!\!\!\! f(x_{k+1}) \!-\! f(x_k) 
\!\!\le\!\! \inp{ \nabla\! f(x_k), x_{k+1} \!-\! x_k } \!+\!\! \frac{L}{2}\| x_{k+1} \!-\! x_{k} \|^2\!. \!\!\!\!
\end{equation}
	
Different from the proof in Theorem \ref{th:vanilla}, we split the RHS of \eqref{apdx04:06} into the sum of the following three terms
\begin{subequations}\label{apdx04:07}
\begin{align}
	& \inp{\nabla f(x_k) - \nabla f(w_k), x_{k+1} - x_k} \label{apdx04:07a}\\
	& \inp{\nabla f(w_k), x_{k+1} - x_k} \label{apdx04:07b}\\
	& \frac{L}{2}\| x_{k+1} - x_{k} \|^2. \label{apdx04:07c}
\end{align}
\end{subequations}

The expectation of \eqref{apdx04:07a} is derived as 
\begin{subequations}\label{apdx04:08}
\begin{align}
& \E{ \inp{\nabla f(x_k) - \nabla f(w_k), x_{k+1} - x_k} } \\
&= \frac{\eta}{1-\beta}\E{ \inp{\nabla f(w_k) - \nabla f(x_k), \frac{1}{N} \sum_{n=1}^N \hat \nabla_{n,k} } } \label{apdx04:08a} \\
&= \frac{\eta}{1-\beta}\E{ \inp{\nabla f(w_k) - \nabla f(x_k), \nabla f(w_k) } } \label{apdx04:08b}\\
%&\le \frac{c_1\eta}{2a(1-\beta)}\E{\|\nabla f(w_k) - \nabla f(x_k)\|^2} \nonumber \\
%&\hspace{0.5 cm} + \frac{a c_1\eta }{2(1-\beta)} \E{ \|\nabla f(w_k)\|^2 }  \label{apdx04:08c}\\
&\le \frac{\eta}{2(1-\beta)}\sq{ \frac{L^2}{a}\E{\|w_k - x_k\|^2} + a \E{ \|\nabla f(w_k)\|^2 } } \label{apdx04:08c} \\
&= \frac{\eta}{2(1-\beta)}\sq{ \frac{\eta^2 L^2 \beta^4}{a(1-\beta)^2}\E{\|u_k\|^2} + a \E{ \|\nabla f(w_k)\|^2 } } \label{apdx04:08d}
\end{align}
\end{subequations}
where the equality \eqref{apdx04:08a} follows from the recursion \eqref{apdx04:02}; 
the equality \eqref{apdx04:08b} is based on the fact $\frac{1}{N} \sum_{n=1}^N \E{ \hat \nabla_{n,k}} = c_1 \nabla f(w_k)$ and $\E{\E{X}} = \E{X}$;
the inequality \eqref{apdx04:08c} follows from the $L$-Lipschitz continuous gradient of loss function and the fact $\inp{x, y} \le \frac{1}{2a}\|x\|^2 + \frac{a}{2}\|y\|^2$ with $a > 0$;
and the equality \eqref{apdx04:08d} is obtained via \eqref{apdx04:03}.

Following similar argument as \eqref{apdx03:03} and recalling \eqref{apdx04:02}, the expectation of \eqref{apdx04:07b} is derived as
\begin{equation}\label{apdx04:09}
\E{ \inp{\nabla f(w_k), x_{k+1} - x_k} } = -\frac{\eta}{1-\beta}\E{ \|\nabla f(w_k)\|^2 }. 
\end{equation}
	
Substituting \eqref{apdx04:02} into \eqref{apdx04:07c}, we obtain 
\begin{equation}\label{apdx04:10}
\EE{ \frac{L}{2}\| x_{k+1} - x_{k} \|^2 } = \frac{\eta^2 L}{2 (1-\beta)^2} \EE{ \| \frac{1}{N} \sum_{n=1}^N \hat \nabla_{n,k} \|^2 }. 
\end{equation}

Summing \eqref{apdx04:08d}, \eqref{apdx04:09}, and \eqref{apdx04:10} and performing several algebraic manipulations, we obtain
\begin{equation}\label{apdx04:11}
\begin{split}
&\E{ f(x_{k+1}) - f(x_k) } \\
&\le \frac{\eta }{1-\beta}\sq{ \frac{a}{2} - 1 }\E{ \|\nabla f(w_k)\|^2 } 
+ \frac{\eta^3 L^2 \beta^4 }{2a(1-\beta)^3} \E{ \|u_k\|^2 }
+ \frac{\eta^2 L}{2 (1-\beta)^2} \EE{ \| \frac{1}{N} \sum_{n=1}^N \hat \nabla_{n,k} \|^2 }.
\end{split}
\end{equation}

Summing \eqref{apdx04:11} over $k = 0, \ldots, K-1$ and recalling \eqref{apdx04:04}, we obtain 
\begin{equation}\label{apdx04:12}
\begin{split}
&\E{ f(x_{K}) - f(x_0) } \\
&\le \frac{\eta }{1-\beta} \sq{ \frac{a}{2} - 1 } \sum_{k=0}^{K-1}\E{ \|\nabla f(w_k)\|^2 } 
+ \sq{ \frac{ \eta^3 L^2  \beta^4 }{2a(1-\beta)^5} + \frac{\eta^2 L }{ 2(1-\beta)^2 } } \sum_{k=0}^{K-1} \EE{ \| \frac{1}{N} \sum_{n=1}^N \hat \nabla_{n,k} \|^2 }. 
\end{split}
\end{equation}

%\begin{figure*}[htb]
%\begin{align}
%\E{ f(x_{k+1}) - f(x_k) } 
%&\le \frac{\eta }{1-\beta}\sq{ \frac{a}{2} - 1 }\E{ \|\nabla f(w_k)\|^2 } 
%	+ \frac{\eta^3 L^2 \beta^4 }{2a(1-\beta)^3} \E{ \|u_k\|^2 }
%	+ \frac{\eta^2 L}{2 (1-\beta)^2} \EE{ \| \frac{1}{N} \sum_{n=1}^N \hat \nabla_{n,k} \|^2 } \label{apdx04:11} \\
%\E{ f(x_{K}) - f(x_0) } 
%&\le \frac{\eta }{1-\beta} \sq{ \frac{a}{2} - 1 } \sum_{k=0}^{K-1}\E{ \|\nabla f(w_k)\|^2 } 
%	+ \sq{ \frac{ \eta^3 L^2  \beta^4 }{2a(1-\beta)^5} + \frac{\eta^2 L }{ 2(1-\beta)^2 } } \sum_{k=0}^{K-1} \EE{ \| \frac{1}{N} \sum_{n=1}^N \hat \nabla_{n,k} \|^2 } \label{apdx04:12}
%\end{align}
%\hrulefill
%\end{figure*}

Setting $a = \frac{ \eta L \beta^3}{(1-\beta)^2}$ and multiplying both sides of  \eqref{apdx04:12} by $\frac{1-\beta}{ \eta }$, we obtain
\begin{subequations}\label{apdx:13}
\begin{align}
&\frac{1-\beta}{ \eta }\E{ f(x_{K}) - f(x_0) }  \label{apdx:13a}\\
&\le \sq{ \frac{ \eta L \beta^3}{2(1-\beta)^2 } - 1 } \sum_{k=0}^{K-1}\E{ \|\nabla f(w_k)\|^2 }  + \frac{ \eta L  }{2 (1-\beta)^2 }  \sum_{k=0}^{K-1} \E{ \| \frac{1}{N} \sum_{n=1}^N \hat \nabla_{n,k} \|^2 } \label{apdx:13b}\\
&\le \sq{ \frac{\eta L(2 + \beta^3)}{2(1-\beta)^2} - 1}\sum_{k=0}^{K-1}\E{ \|\nabla f(w_k)\|^2 } \ \frac{c_1 \!+\! c_2 d_1 p_0^{-1} \sigma^2 \!+\! G}{4(1-\beta)^2 N}\eta L K  \label{apdx:13c}
\end{align}
\end{subequations}
where \eqref{apdx:13c} follows from Lemma \ref{le:02}. 
	
Setting $\eta \le \frac{3(1-\beta)^2}{ 2(2 + \beta^3) L}$, we have $1 - \frac{\eta L(2 +  \beta^3)}{2   (1-\beta)^2} \ge \frac{1}{4}$. 
Dividing both sides of \eqref{apdx:13c} by $K$ and using $f(x_K) \ge f^*$, we obtain
\begin{align}
&\frac{1}{K}  \sum_{k=0}^{K-1} \E{ \| \nabla f(w_k) \|^2 } \label{apdx:14}\\
%		&\le \frac{ 1-\beta }{\eta c_1 K} \E{ f(x_{0}) - f(x_K) } \\
%		&\hspace{0.5 cm} + \frac{\eta L (c_1^{-1} - 1)}{ (1-\beta)^2} G + \frac{\eta L c_2 d_1 \sigma^2 }{2 p_0 N (1-\beta)^2} G\\
&\le \frac{ 4(1 \!-\! \beta) }{\eta K} \E{ f(x_{0}) \!-\! f^* }  \!+\! \frac{c_1 \!+\! c_2 d_1 p_0^{-1} \sigma^2 \!+\! G  }{ (1-\beta)^2 N}\eta L. \nonumber
\end{align}

we complete the proof. 
	
\bibliographystyle{IEEEtran}
\bibliography{dyj_bib}

\begin{IEEEbiographynophoto}{Yanjie Dong (Member, IEEE)} is with Shenzhen MSU-BIT university as an Associate Professor. 
	Dr. Dong obtained his PhD and MASc degree from The University of British Columbia, Canada, in 2020 and 2016, respectively. His research interests focus on the protocol design of energy-efficient communications, machine learning based resource allocation algorithms, and quantum computing technologies. 
	He regularly serves as a member of Technical Program Committee in flagship conferences in IEEE ComSoc.
	%	He was awarded a Four-Year Doctoral Fellowship of UBC in 2016. He received the Graduate Support Initiative Award in 2016, 2017, and 2018, the British Columbia Graduate Scholarship in 2019, the MITACS Globalink Research Award in 2019, Chinese Government Award for Outstanding Self-financed Students Abroad in 2019.
\end{IEEEbiographynophoto}

\begin{IEEEbiographynophoto}{Luya Wang}
	received his Bachelor's degree from Shandong University of Science and Technology in 2020. He is pursuing a Master's degree in the School of Computer and Software at Shenzhen University, and is also a visiting student of Shenzhen MSU-BIT University, Shenzhen, China. His research interests focus on machine learning, federation learning, and wireless communications.
\end{IEEEbiographynophoto}

\begin{IEEEbiographynophoto}{Yuanfang Chi (Student Member, IEEE)}
	received the BASc. and MASc degrees from The University of British Columbia, Vancouver, BC, Canada, in 2012 and 2015, respectively, where she is currently pursuing the Ph.D. degree with the Department of Electrical and Computer Engineering.
	She is a visiting student of College of Computer Science and Software Engineering, Shenzhen University, Shenzhen, China. Her current research interests are distributed machine learning, knowledge graph, fault diagnosis, and Industrial Internet of Things.
\end{IEEEbiographynophoto}

\begin{IEEEbiographynophoto}{Jia Wang (Member, IEEE)}
	received her PhD degree in Electronic Engineering from City University of Hong Kong, Hong Kong SAR of China, in 2017. She is now an Assistant Professor with the College of Computer and Software Engineering, Shenzhen University. Her current research interests include lightweight security design for IoT and data security and privacy-preserving aspects of machine learning. 
\end{IEEEbiographynophoto}

\begin{IEEEbiographynophoto}{Haijun Zhang (Fellow, IEEE)}
	is currently a Full Professor and Associate Dean in School of Computer and Communications Engineering at University of Science and Technology Beijing, China. He was a Postdoctoral Research Fellow in Department of Electrical and Computer Engineering, the University of British Columbia, Canada. 
	He serves as an Editor of IEEE Transactions on Communications, and IEEE Transactions on Network Science and Engineering. He received the IEEE CSIM Technical Committee Best Journal Paper Award in 2018, IEEE ComSoc Young Author Best Paper Award in 2017, and IEEE ComSoc Asia-Pacific Best Young Researcher Award in 2019. 
\end{IEEEbiographynophoto}

\begin{IEEEbiographynophoto}{F. Richard Yu (Fellow, IEEE)}
	received the PhD degree in electrical engineering from the University of British Columbia in 2003. His research interests include connected/autonomous vehicles, artificial intelligence, cybersecurity, and wireless systems. He has been named in the Clarivate Analytics list of ``Highly Cited Researchers" since 2019, and received several Best Paper Awards from some first-tier conferences. He is an elected member of the Board of Governors of the IEEE VTS and Editor-in-Chief for IEEE VTS Mobile World newsletter. He is a Fellow of the IEEE, Canadian Academy of Engineering, Engineering Institute of Canada, and IET. He is a Distinguished Lecturer of IEEE in VTS and ComSoc.
\end{IEEEbiographynophoto}

\begin{IEEEbiographynophoto}{Victor C. M. Leung (Life Fellow, IEEE)}
	is a Distinguished Professor of Computer Science and Software Engineering at Shenzhen University, China. He is also an Emeritus Professor of Electrical and Computer Engineering and Director of the Laboratory for Wireless Networks and Mobile Systems at the University of British Columbia (UBC), Canada.  His research is in the broad areas of wireless networks and mobile systems, and he has published widely in these areas. Dr. Leung is serving as a Senior Editor of the IEEE Transactions on Green Communications and Networking. He is also serving on the editorial boards of the IEEE Transactions on Cloud Computing, IEEE Transactions on Computational Social Systems, IEEE Access, IEEE Network, and several other journals. He received the 1977 APEBC Gold Medal, 1977-1981 NSERC Postgraduate Scholarships, IEEE Vancouver Section Centennial Award, 2011 UBC Killam Research Prize, 2017 Canadian Award for Telecommunications Research, 2018 IEEE TCGCC Distinguished Technical Achievement Recognition Award, and 2018 ACM MSWiM Reginald Fessenden Award. He co-authored papers that won the 2017 IEEE ComSoc Fred W. Ellersick Prize, 2017 IEEE Systems Journal Best Paper Award, 2018 IEEE CSIM Best Journal Paper Award, and 2019 IEEE TCGCC Best Journal Paper Award.  He is a Life Fellow of IEEE, and a Fellow of the Royal Society of Canada (Academy of Science), Canadian Academy of Engineering, and Engineering Institute of Canada. He is named in the current Clarivate Analytics list of ``Highly Cited Researchers''.
\end{IEEEbiographynophoto}

\begin{IEEEbiographynophoto}{Xiping Hu (Member, IEEE)}  is currently a Distinguished Professor with Shenzhen MSU-BIT University, and is also with Beijing Institute of Technology, China. Dr. Hu received the Ph.D. degree from the University of British Columbia, Vancouver, BC, Canada. Dr. Hu is the co-founder and chief scientist of Erudite Education Group Limited, Hong Kong, a leading language learning mobile application company with over 100 million users, and listed as top 2 language education platform globally. His research interests include affective computing, mobile cyber-physical systems, crowdsensing, social networks, and cloud computing. He has published more than 150 papers  in the prestigious conferences and journals, such as IJCAI, AAAI, ACM MobiCom, WWW, and IEEE TPAMI/TIP/TPDS/TMC/COMMAG.
\end{IEEEbiographynophoto}

\end{document}